\documentclass[english]{article}
\usepackage{mathptmx}
\usepackage{helvet}
\usepackage{courier}
\usepackage[T1]{fontenc}
\usepackage[latin9]{inputenc}
\usepackage{color}
\usepackage{array}
\usepackage{rotating}
\usepackage{float}
\usepackage{url}
\usepackage{multirow}
\usepackage{amsmath}
\usepackage{amssymb}
\usepackage{graphicx}

\makeatletter

\newcommand{\lyxmathsym}[1]{\ifmmode\begingroup\def\b@ld{bold}
  \text{\ifx\math@version\b@ld\bfseries\fi#1}\endgroup\else#1\fi}

\providecommand{\tabularnewline}{\\}

\makeatother

\usepackage{babel}
\begin{document}

\title{Durkheim Project Data Analysis Report }

\date{23 October 2013 }

\author{Linas Vepstas }
\maketitle
\begin{abstract}
This report describes the suicidality prediction models created under
the DARPA DCAPS program\cite{Poulin2013}. The models were built primarily
from unstructured text (free-format clinician notes) for several hundred
patient records obtained from the Veterans Health Administration (VHA).
The models were constructed using a genetic programming algorithm
applied to bag-of-words and bag-of-phrases datasets. The influence
of additional structured data was explored but was found to be minor.
Given the small dataset size, classification between cohorts was high
fidelity (98\%). Cross-validation suggests these models are reasonably
predictive, with an accuracy of 50\% to 69\% on five rotating folds,
with ensemble averages of 58\% to 67\%. One particularly noteworthy
result is that word-pairs can dramatically improve classification
accuracy; but this is the case only when one of the words in the pair
is already known to have a high predictive value. By contrast, the
set of all possible word-pairs does not improve on a simple bag-of-words
model.
\end{abstract}

\section*{Introduction}

A central goal of the Durkheim Project (\url{http://durkheimproject.org/})
is to build a classifier for suicide ideation and prediction of suicide
risk, based on free-text clinician notes contained in a set medical
records obtained from the Veterans Health Administration (VHA), via
the Dartmouth-Hitchcock Medical Center (DHMC). The intended use of
the classifier is to aid the clinician in determining the suicide
risk of prospective patients. As such, it should be able to digest
patient data, and assign a risk level, green/yellow/red, suggesting
the likelihood of suicidal ideation.

In order to understand how to build such a classifier, an extensive
analysis of medical records of VHA patients were performed. Patient
records were divided into three cohorts. These consist of a control
group of 70 patients (group 1), a suicide cohort of 69 patients (group
2), and a psychiatric cohort of 70 patients (group 3). The medical
records consist primarily of free-text notes entered by the clinician,
as well as additional structured data (demographics, drug prescriptions,
hospitalization admission records). The inclusion of the structured
data in the training set makes for a very slight improvement of the
overall score (fractions of a percent).

The clinician notes include both nurse and doctor notes, ranging from
mundane procedures '\emph{Patient received influenza vaccine per order}',
descriptions of more serious procedures: '\emph{ultrasound of the
abdominal aorta done on...}', a number of semi-automatic script-generated
tables: '\emph{Issue Date Status Last Fill Outpatient Medications
(By Class) Refills Expiration}', as well as psychologically charged
entries: '\emph{Little interest or pleasure in doing things}'. Notes
that discuss psychological state, including screenings for depression
and alcoholism, appear in all three cohorts. One may presume that
these are far more common, and delve deeper, in the last two cohorts.
Aside from this kind of quick, cursory review to validate the general
form of the records, no deeper review or examination was performed.

The data analysis was performed by using supervised training with
a genetic programming system to build models of the datasets. The
models were constructed by converting the free-text records into a
'bag of words': a simple numerical count of how often a given word
appears in the context of a certain patient record. Any given model
then identifies which words, taken in combination, serve as predictors
of suicide. The nature of the genetic programming system used is that
it can build many different models, depending on an initial random
seed. Thus, data analysis consisted primarily of generating ensemble
averages of models trained on the same dataset. Model validation was
performed by using 5-fold cross-validation: that is, by setting aside
1/5 of the dataset for testing, and training on the remaining 4/5ths.
Model accuracy was used as the score: that is, the total fraction
of correct answers.

Most of the data analysis was focused on building a binary classifier
to distinguish group 2 and 3. This was done for several reasons. One
important reason was simply that these were the two largest groups,
in terms of total word-count, and thus presented the greatest amount
of data to work with. An equally important reason, though, is the
clinical perception that these two groups are hard or even impossible
to distinguish. By contrast, the control group consists of patients
obtaining non-psychiatric medical care, and thus is almost completely
devoid of references to psychological state. As such, it should be,
in principle, easy to distinguish simply because it lacks this vocabulary.
Results for binary classifiers trained to distinguish groups 1 vs.
2, as well as group 1+3 vs. group 2, are also presented.

Many of the words appearing in the models are emotionally charged
or psychologically significant, such as 'PTSD', 'weapons', or 'overdose'.
Taken individually, these words are meaningful, but not clinically
out of the ordinary. Thus, a question arises: what phrases are these
words a part of? Thus, a phrase such as 'negative assessment for PTSD'
carries a different meaning than 'positive assessment', and is thus
a potentially useful feature for classifying suicidal patients. This
suggests that a 'bag-of-phrases' approach may be more accurate than
a bag-of-words model, and this was indeed found to be the case. In
particular, models built using certain word-pairs had significantly
better scores than the single-word models, and had the best scores
overall. Besides word pairs (bi-grams), trigrams and 4-grams were
also explored, but these did not offer improvements, and it is hypothesized
that the training datasets were too small to have a noticeable effect
for these. The improvement for word-pairs is seen only when an initial
selection or 'cut' is made: a word-pair is used only if one of the
words in the pair already correlates well with the cohort. Without
this cut, using word-pairs does not improve the score, and in fact
lowers it: it is easier to over-train in such a case.

For single-word bag-of-word models, the accuracy, averaged over 100
models, was typically seen to be about 58\% (depending on which cohorts
were being distinguished), with low and high scores for individual
models ranging from 46\% (worse than random chance) to over 65\%.
For selectively-chosen word-pairs, individual model scores ranged
from 52\% to 69\%, with an ensemble average (for 100 models) of 67\%.
To the authors, this appears to be a remarkable achievement, given
the small size of the dataset and the fragmentary nature of clinician
notes.

The remainder of this document is structured to provide a mode detailed
review of the model building and validation process, the size and
content of the clinician notes, and the various results obtained.

\section*{Model Building and Validation}

Model building consists of several stages. The initial stage converts
the free-text data into a ``bag of words''. This is simply a count
of word frequency, and nothing more: a count of how often some given
word was used in a particular patient's medical report. Bag-of-words
models completely ignore any sort of linguistic structure in the original
text, as well as ignoring punctuation and any structural markup (paragraphs,
sentence endings, \emph{etc}.). Typically, 30 to 40 thousand different
words were found, depending on which cohort is examined. These words
were not spell-checked nor stemmed, and include many typographical
errors as well as a large number of abbreviations for hospitals, clinics,
departments, tests, procedures, and orders.

The next stage consists of 'feature selection'. Rather than training
the discriminator directly on the full set of word counts, the set
is reduced to the several thousand words judged to be most significant
in predicting outcome. The cut may be done in several ways. One possible
cut is to remove words that occur less than a few dozen times. Although
the intent of this cut is to remove noise from the data, it is possible
that perhaps some significant indicators are lost as well; thus data
analysis includes experiments adjusting this cut. Another possible
cut is to only count word stems: that is, to consolidate the counts
for singular and plural forms of a noun, and to consolidate past,
present and future tenses of verbs. The most important cut is to choose
only those words whose counts correlate well with the patient grouping.
This is done by computing the 'mutual information' (MI) between the
group id (1, 2 or 3) and the word-count frequency. The few thousand
words with the highest MI are then selected to be used for the final
model-building stage.

Feature selection is an important step of model building, and has
a counter-intuitive effect on the final model: it is often the case
that limiting the number of features used to build the model results
in a better, more accurate model. This is because machine-learning
algorithms can often focus in on irrelevant differences when classifying
into groups: the differences are irrelevant, in that they fail to
have predictive value. The greater the number of features (words)
given to such a learning algorithm, the more likely it is to find
such irrelevant differences; limiting the input to only the most significant
features helps prevent such over-training.

Model building was performed using the poses/moses machine learning
system\cite{Looks2007,Looks2006}. This system builds candidate representative
models or 'representations' of the data, and then uses evolutionary
algorithms to discover the most effective representation. An example
of such a representation, one of many, trained on the current data,
is shown in Table \ref{tab:Example-Representation}.

\begin{table}[h]
\caption{Example Representation\label{tab:Example-Representation}}
\vphantom{}

~~~~%
\begin{minipage}[t]{0.95\columnwidth}%
\begin{flushleft}
\texttt{or(and(or(and(\$MODERATE\_t1.3 !\$PRESCRIBE\_t0.02) \$CONCERN\_t0.8
\$EVIDENCE\_t0.4 \$INCREASING\_t0.3 \$RESTRICTED\_t0.1) or(\$ALBUTEROL\_t1.2
\$AMOUNTS\_t0.08 \$SYSTEM\_t0.08 \$VIEW\_t0.8) or(!\$STOMACH\_t0.4
!\$SURROGATE\_t0.7)) and(!\$BRING\_t0.6 !\$HIGH\_t1.9 !\$MINUTES\_t2.5
!\$SAT\_t0.7 \$STOMACH\_t0.4) \$LOWEST\_t0.08 \$NYSTAGMUS\_t0.03 \$OLANZAPINE\_t0.05
\$OVERDOSE\_t0.09 \$PRESCRIBE\_t0.02 \$SUPERFICIAL\_t0.16 \$WEAPONS\_t0.04
\$WITHDRAWAL\_t0.2)}
\par\end{flushleft}%
\end{minipage}\\
\vphantom{}

The above is an example of a representation built from the VHA dataset.
It may be understood as follows: \texttt{\$MODERATE\_t1.3} takes on
a value of 'true' if the word 'moderate' occurs 1.3 or more times
in the text (floating point values are used in case word-counts have
been normalized to non-integer values). The exclamation mark ! indicates
that the condition does not hold: so \texttt{!\$PRESCRIBE\_t0.02}
means that the word 'prescribe' does NOT occur 0.02 or more times.
The Boolean operators 'and', 'or' serve to conjoin these conditions:
thus the above is saying that, ``if the word 'moderate' appears at
least twice, and the word 'prescribe' does not appear, or if any of
the words 'concern', 'evidence', 'increasing' or 'restricted' appear
at least once, and the word 'albuterol' appears at least twice ...
then the patient should be classified as belonging to group 2.''
Note that, out of the approximately twenty-five thousand unique words
appearing in the data, the above is really a rather very small subset.

\begin{minipage}[t]{1\columnwidth}%
\rule[0.5ex]{1\columnwidth}{1pt}%
\end{minipage}
\end{table}

The final classifier consists of not just one such representation,
but many, ranging from one to over a hundred, depending on parameter
settings. The predictions of each representative is used to cast a
vote; the final determination follows from a tally of these votes.
This process of ensemble averaging eliminates a considerable variation
of accuracy from one model to the next\cite{Opitz1999}.

To determine the accuracy and performance of the classifier, standard
k-fold cross-validation techniques are used, with k=5. In this style
of validation, the dataset is divided into 5 parts. Four of the parts
are used to train a model, and then the accuracy of the model is measured
on the fifth part. One then repeats this process, each time leaving
out a different fifth of the dataset, to be used for evaluation. The
average of the five sessions may then be given as the overall accuracy.

Almost all of the data analysis reported here was done by training
the classifier to maximize accuracy: that is, to minimize the sum
of the false-positive and false-negative rates. This is the appropriate
approach when the datasets are balanced in size, as they are here.
Alternatives to maximizing the accuracy would be maximizing the $F_{1}$-score
or $F_{2}$-score, maximizing the recall rate, or the precision. None
of these alternatives seem particularly suited for this dataset; they
can lead to unexpected, imbalanced effects. For example, it will be
seen later that it appears to be considerably easier to pick out patients
with a low suicide risk out of a mixed population, than to pick out
those with a high risk; this is covered in a later section. However,
for a general population wherethe suicide risk is very low, such ideas
would need to be re-examined.

The remained of this document expands on each of the steps above in
greater detail, describing feature selection, model building, and
the estimation of the accuracy of the models.

\section*{Dataset statistics}

The data consists of three sets of medical records:
\begin{itemize}
\item Group 1: The control cohort. These are the records of 70 patients
who sought medical attention, but did not require or receive any special
psychiatric treatment.
\item Group 2: The suicide cohort. These are the records of 69 patients
that committed suicide.
\item Group 3: The psychiatric control group. These are records of 70 patients
who sought help with psychiatric issues; they have not committed suicide,
but may be at risk.
\end{itemize}
Associated with each patient is a set of note records, covering the
span of one year. Records are generated for many reasons: upon hospital
or clinic intake (by nursing staff); patient care notes (by the primary
physician); examination results; lab results; consultation notes;
notes from referrals, including imaging; outpatient notes; surgery
and treatment notes; pharmacy notes; ongoing therapy notes; telephone
follow-up notes; addenda and corrections. Thus, a single patient visit
on a single day can generate from one to more than a dozen records. 

The dataset is tokenized into a bag of words by converting all punctuation
into white-space, and using white-space as word separators. The exceptions
were word-phrases that included hyphens or underscores; this punctuation
was simply removed to create a single run-on word. Differences in
capitalization were ignored by converting all words to upper-case.
After this normalization, the dataset was found to consist of nearly
one million words; precisely, 971,189 words total. These are distributed
across the three groups as follows: 
\begin{itemize}
\item Group 1: 155,354 words, or 2,219 words per patient.
\item Group 2: 350,435 words, or 5,079 words per patient.
\item Group 3: 465,400 words, or 6,648 words per patient.
\end{itemize}
The number of words per record is fairly uniform across all three
cohorts. Record lengths were limited to 1024 characters per record;
it is clear that some of the longer records were truncated mid-sentence,
mid-word. This appears to be due to technical interoperability difficulties
with the VA data processing systems. 
\begin{itemize}
\item Group 1: 1,913 records, or 27 records per patient, 81 words per record.
\item Group 2: 4,243 records, or 61 records per patient, 82 words per record.
\item Group 3: 5,388 records, or 77 records per patient, 86 words per record.
\end{itemize}
There were 24,860 unique words in the dataset that occurred at least
once, but only 14,728 that occurred twice or more. A rough sketch
of the distribution is given in table \ref{tab:Word-Distribution}.
Many of the words that appear only once are typos and miss-spellings
of common words, abbreviations of medical terms, and a fair number
of acronyms, including abbreviated names of clinics and hospital departments,
lab procedures, orders and prescriptions. However, there are also
many non-misspelled words that appear only once in the text, such
as: ABANDONMENT ABORTIVE ABORTED ABUSER ABUSES ABYSS ACADEMY ACCUSE
ACHIEVABLE ACHIEVES ACQUAINTED. Note that many of these words are
emotionally meaningful words. Whether these infrequently-used can
serve as indicators of psychological state is unclear. Experiments
where low-frequency words are removed from the dataset before model
building are reported below. At any rate, it is clear that the 'active
vocabulary' of frequently used words is fairly small.

There was no attempt made to extract word stems, nor to correct or
exclude 'obvious' miss-spellings. Whether doing so would enhance or
diminish the ability to categorize is not clear \emph{a priori}. No
inclusion or exclusion criteria based on vocabulary were applied.
Many different cuts, based on word-counts and mutual information,
were explored, as detailed below. A feature selection stage applied
prior to model building also effectively removes the majority of words
from further consideration, but this cut is based purely on the predictive
utility of a word, and not on its morphology, spelling, lexical meaning
or usage. 

\begin{table}[h]
\caption{Word Distribution\label{tab:Word-Distribution}}

\vphantom{}

\begin{centering}
\begin{tabular}{|c|c|}
\hline 
Count & Number of occurrences\tabularnewline
\hline 
\hline 
24861 & once or more\tabularnewline
\hline 
14728 & 2 times or more\tabularnewline
\hline 
11613 & 3 or more\tabularnewline
\hline 
9928 & 4 or more\tabularnewline
\hline 
8844 & 5 or more\tabularnewline
\hline 
6862 & 8 or more\tabularnewline
\hline 
4618 & 16 or more\tabularnewline
\hline 
3042 & 32 or more\tabularnewline
\hline 
1928 & 64 or more\tabularnewline
\hline 
\end{tabular}
\par\end{centering}

\vphantom{}

\begin{centering}
Dataset word distribution.
\par\end{centering}

\begin{minipage}[t]{1\columnwidth}%
\rule[0.5ex]{1\columnwidth}{1pt}%
\end{minipage}
\end{table}

The most frequently occurring words are shown in table \ref{tab:Most-Frequent-Words}.
Function words (the so-called 'stop' words) were not removed from
the dataset, and thus appear in this table. There is a good reason
for this: function words are known to be strong indicators of psychological
state, and, in particular, the writing of suicides is known to make
greater use of function words and pronouns than average\cite{Penne2001,Penne2007}. 

\begin{table}[h]
\caption{Most Frequent Words\label{tab:Most-Frequent-Words}}

\vphantom{}

\begin{centering}
\begin{tabular}{|c|c|}
\hline 
Word & $\log_{2}$ probability\tabularnewline
\hline 
\hline 
TO & -5.191\tabularnewline
\hline 
AND & -5.542\tabularnewline
\hline 
THE & -5.568\tabularnewline
\hline 
OF & -5.755\tabularnewline
\hline 
FOR & -6.124\tabularnewline
\hline 
PATIENT & -6.151\tabularnewline
\hline 
HE & -6.418\tabularnewline
\hline 
\end{tabular} \\
\vphantom{}
\par\end{centering}

The probability of a word is obtained by taking the number of times
the word occurs, and dividing by the total word count. Here, $\log_{2}$
denotes the logarithm base-2. Thus, ``to'' occurs 26,588 times,
or $0.027=2^{-5.191}$ fraction of the time. 

\begin{minipage}[t]{1\columnwidth}%
\rule[0.5ex]{1\columnwidth}{1pt}%
\end{minipage}
\end{table}

\begin{figure}
\caption{Word Rank Distribution\label{fig:Word-Rank-Distribution}}

\includegraphics[width=0.95\columnwidth]{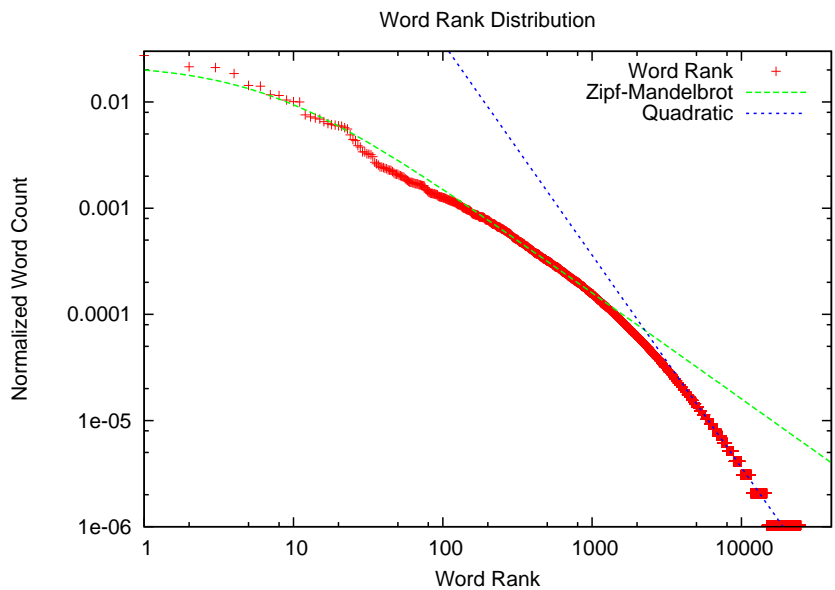}

Word rank distribution for the dataset. The 'normalized word count'
is the frequency with which a word appears in the dataset. The 'rank'
is order of a word, when sorted by frequency. The green line indicates
the Zipf-Mandelbrot law, here given as $0.16\times(\mbox{rank}+7)^{-1}$.
The blue line is a quadratic fit, given by $360\times(\mbox{rank})^{-2}$.
The word distribution for most English-language texts (books, newspapers)
is much flatter than that shown here. When word-pairs are incorporated
into this ranking the curve also flattens and becomes less steep.

\begin{minipage}[t]{1\columnwidth}%
\rule[0.5ex]{1\columnwidth}{1pt}%
\end{minipage}
\end{figure}

The overall word distribution appears to obey the Zipf-Mandelbrot
law (modified Zipf law), with a quadratic fall-off tail. This is more
curved, and with a more quickly falling tail, than is commonly the
case for natural-language texts. The distribution is shown in fig
\ref{fig:Word-Rank-Distribution}.

Word-pairs were also explored, as these have a predictive power as
well. Word pairs were constructed by considering adjacent words, as
well as pairs one word apart (ignoring the word in the middle). Thus,
for example: \textquotedbl{}big red balloon\textquotedbl{} generates
three word pairs: \textquotedbl{}big\_red\textquotedbl{}, \textquotedbl{}red\_balloon\textquotedbl{}
and ``big\_balloon''. The first of these pairs is not particularly
meaningful, but both of the last two are semantic units. The last,
``big\_balloon'', would not have been captured if one confined oneself
only to adjacent words. By eliding middle words such semantically
significant pairs can be discovered. 

Not all word pairs are equally interesting. Semantically meaningful
word pairs are those with a high mutual information between them.
Mutual information (MI) for a pair of words x,y is defined as 
\[
MI(x,y)=-\log_{2}\frac{p(x,y)}{p(x,*)p(*,y)}
\]
Here, $p(x,y)$ is the probability of seeing the word pair x,y, divided
by the total number of word pairs. The two probabilities $p(x,*)$
and $p(*,y)$ are the probabilities of seeing any word pair, whose
first word is $x$, or last word is $y$, respectively. In general,
MI scores typically range from slightly above 20 to less than zero;
the same is true of this dataset. In general, word pairs with a high
MI form lexical units, conveying meaning, that is, having semantic
content. They are collocations, often forming idioms and set phrases.
Examples of word pairs with an MI of about 20, taken from this dataset,
include ULTERIOR\_MOTIVES, HLTHY\_LVNG, VOCALIZES\_INTELLIGIBELY,
GIN\_TONICS, ROAST\_BEEF, MARATHON\_RUNNER, GOVERNMENTAL\_ENTITIES.
By contrast, lower MI scores are less meaningful. Typically, the boundary
between meaningful and meaningless word pairs occurs around an MI
of 2 to 4. Examples of MI of 4 from this dataset include: HUNGRY\_HAD,
HAD\_SWEAT, INTERACT\_IN, RX\_IBUPROFEN, ANYTHING\_HIMSELF. Those
with an MI below zero degenerate into random nonsense: MORPHINE\_YOU,
RECOVERY\_ARE, HIS\_HOW, YES\_WITH: pairs of words that appear next
to one-another purely by coincidence, and not due to any linguistic
construction. Thus, the mutual information can be used as a cut, to
exclude low-MI word pairs from consideration during model building.

Results from models built from a variety of different MI cuts are
presented below. Word pairs can be ranked along with individual words;
the overall shape of the distribution does not change much; it is
similar to that shown in figure \ref{fig:Word-Rank-Distribution},
but considerably flatter, loosing the quadratic fall-off for low frequency
words.

\section*{Model Building and Validation Details}

The various stages of feature selection, model building and validation
are each in themselves rather complex, and require some care to perform
properly. None of the stages are 'pre-determined' or 'automatic';
instead, each has adjustable parameters and requires a deliberate
choice of these parameters and overall configuration. Since the accuracy
of the final classifiers depends on the various parameters settings
in the data processing stages, it is important to understand what
these are and how they are applied. The sections immediately below
provide details describing these stages. This is followed by a presentation
of the results obtained as these stages are applied.

\subsection*{Binning}

Prior to performing training on the dataset, bin-counts are created.
Binning helps to make up for relatively sparse data by lumping together
similar word-counts into the same category or 'bin'. This serves to
further simplify the data and boost the performance of the training
step. It is performed by counting how often a word occurs for a given
patient, and assigning it to a bin, such as 'occurs more than twice,
but less than four times'. For a fixed set of bins, different patient
records will be seen to contain different numbers of words in them.

A set of natural bin sizes can be obtained by first determining the
probability distribution of a given word (over all patients); that
is, by determining the average number of times it occurs (across all
patients), and the standard deviation about this average (as it varies
from patient to patient). These two numbers provide a natural size
for a bin. For example, given an average number of times that a word
occurs in a patient record, one may then say that, for a given patient,
a given word occurs more than average, or less than average; in this
case, there are two bins total. Another possibility is to use three
bins: for a given patient, a word may occur about an average number
of times (to within one standard deviation away from this average),
or well-below average (more than one standard deviation below average),
or well above average (more than one standard deviation above average).
Bins serve to 'smooth' and consolidate word counts and make them more
granular, to 'filter out high-frequency noise' from the data. In general,
the less data one has, the fewer bins should be used, thus keeping
the bins fairly full. Two to five bins may be considered; it will
be seen, in later sections, that two bins work best for this dataset.

The result of binning are Boolean-valued features. So, for example,
if the term 'PTSD' occurs an average of 2 times per patient record,
a two-bin system would create one feature for this word: $(PTSD>2)$
which is either true or false for a given patient record. If, for
example, the standard deviation was 1.0 for this word, a three-bin
system would include two features for this word, set at one standard
deviation above and below average; that is, $(PTSD>1)$ and $(PTSD>3)$,
each of which may be true or false for any given patient record. The
values '2', '1', '3' shown here are referred to as 'thresholds': they
are the boundaries between the bins. Thus, specifying N thresholds
results in N+1 bins. 

The number of thresholds to use is a parameter that can be specified;
varying this parameter results in models of varying accuracy. The
number of thresholds used is the same for all word counts: thus, setting
thresholds=1 specifies that two bins are to be used for \emph{all}
words. So, for example, given 31 thousand distinct words, a two-bin
system would create 31 thousand (true-false) features, while a three-bin
system would result in twice as many: 62 thousand Boolean-valued features.
A four-bin system would result in three times as many features, and
so on. As is clear, increasing the number of thresholds vastly increases
the dimensionality of the feature space.

\subsection*{Feature Selection}

After binning, but before building a model, the dataset, now converted
into a collection for true/false bin assignments, is run through a
static feature-selection stage. This is done to reduce the size of
the dataset, from tens of thousands of features, to a few thousand.
The goal of this reduction is simply to improve the run-time and memory
usage of the model-building stage.

Given that the overall dataset consists of only a few hundred records,
it may seem reasonable that at most a few hundred features would suffice
to provide predictive value; and indeed, the final models consist
of dozens of words. However, the run-time speed of the next stage,
model-building, is not strongly affected by the number of features
that it is given, and so it was deemed safer to err on the side of
giving it too many features to choose from (thousands), rather than
too few (hundreds). Because of this, a very simple and efficient feature
selection algorithm suffices. The algorithm used is to choose those
features that have the highest mutual information with the desired
patient classification. The mutual information is defined in the same
way as before: 
\[
MI(x,y)=-\log_{2}\frac{p(x,y)}{p(x,*)p(*,y)}
\]
except that here, the variable $x$ is taken as the classification
of a patient belonging to one group or another, while the variable
$y$ is taken to denote whether a given feature is true or false.
Thus, if a certain feature is true whenever the patient belongs to
group A, we expect $MI(A,true)$ to be large; likewise, it may anti-correlate:
$MI(A,false)$ may be large. To continue with the previous example,
the mutual information content $MI(\lyxmathsym{\textquotedblleft}group\,1\lyxmathsym{\textquotedblright},\,\,(PTSD>2))$
is computed for the word 'PTSD'. If this MI ranks in the top few thousand,
then $(PTSD>2)$ is accepted as a valid feature worth exploring during
the training stage. Words that occur equally often in one group as
another will have a low MI score, and thus will not be selected. In
most of the data analysis presented below, the highest-ranked 3000
features were selected. This represents anywhere from 2\% to 10\%
of the total number of features, depending on the number of bins chosen,
and the particular datasets examined.

There is no particularly strong reason for choosing MI, as opposed
to some other correlation measure, such as tf-idf. MI has a strong
mathematical foundation rooted in maximum entropy principles. It does
not discriminate against rare words; a word which occurs infrequently
but still correlates well with the patient grouping will have a reasonable
MI score, and thus will be eminently suitable for use in a classifier.
In any case, the particular choice of a feature selection algorithm
should have little impact on model building.

\subsection*{Model Building}

The most technically difficult and CPU intensive stage of the processing
is the creation of models of the data. This step is performed by the
Meta-Optimizing Semantic Evolutionary Search (MOSES) system. This
system searches through a very large set of representations, such
as that shown in table \ref{tab:Example-Representation}, and locates
those that most accurately fit the training data. 

The MOSES algorithm consists of two nested loops: representation-building
and genetic-algorithm search\cite{Looks2007,Looks2006}. The system
starts by creating a program tree (in the current usage, a tree of
Boolean operators, such as that shown in table \ref{tab:Example-Representation}).
The nodes and leaves of the tree are free to vary over the full range
of input variables, as well as to vary over the Boolean operators
(\emph{and, or, not}). For any fixed choice of nodes and leaves, the
resulting tree may be scored against the input training data (the
features) to see how well it fits; clearly some choices will be better
than others. The set of node and leaf settings are explored using
a genetic evolutionary search algorithm combining hill-climbing and
genetic cross-over. When no further improvements are found, the process
is begun again, this time with a different, and usually, a more complex
program tree. This step is again repeated until either a perfect score
is reached, or set time-limits are exceeded. 

The generation of candidate program trees involves a second, 'dynamic'
feature-selection stage. A new candidate tree is created from an older
high-scoring tree, by decorating it with additional candidate features.
Rather than creating a candidate program tree with all of the several
thousand features in it, convergence can be improved by working only
with those features that add new information to those that are already
in the tree: that is, by working with those features most likely to
improve the current high-scoring tree. This is again a form of feature-selection,
hereinafter referred to as 'dynamic feature selection', as the selected
features depend on the program tree as well as the dataset, and a
different set is chosen for each program tree. Training can be effective
even with a very small number of dynamically selected features: best
results are achieved with less than one hundred, and the technique
is highly effective with as little as five! Aside from improving scores,
working with a smaller number of features dramatically reduces training
time.

The result of this process is a large number of representations, each
of which model the training data more or less equally well. Each representation
may be used to classify new patients (patients not in the training
set); that is, to make predictions about the value of the dependent
variable (the patient classification) based on input variables (word
counts). When tested on a test set of patients held out from the training
group, it can be seen that the accuracy of these representations on
the test data is considerably variable. There is no \emph{a priori}
way of knowing which representation performs 'the best' on the test
data. To overcome this variability, an \emph{ensemble} is created,
with each representation in the ensemble getting a vote to determine
the final classification. That is, the same inputs are presented to
each representation, with each representation making a prediction:
a majority vote is then taken to determine the final classification.
This ensemble is referred to as the \emph{model}, as it is effectively
a distilled, compressed version of the training data.

The theoretical validity of using the model for classification in
this way is founded on the belief that the model captures something
essential about the way that words are used in text. This is a reasonable
belief, given industry experience with bag-of-words classifiers. The
practical validity of the model can be tested in several ways; \emph{k}-fold
cross-validation will be used here.

\subsection*{Cross-Validation}

In order to test the validity of the models, \emph{k}-fold cross-validation
is performed, with $k=5$. The input dataset is split into \emph{k}
subsets, with each subset containing $1/k$ of the patient records,
assigned by round-robin selection. Training is then performed using
$k-1$ of these subsets as input, and a model is built (that is, a
model is build on 4/5'ths of the data). The accuracy of the model
is then evaluated on the subset that was held out (on the remaining
1/5th). This process is repeated $k$ times, to obtain $k$ models,
and $k$ different accuracy test results. The test results are then
averaged together to obtain an estimate to the overall system accuracy.
That is, if a model were trained on the full data-set (without any
hold-outs), the accuracy of this resulting model, on new, blind data,
is expected to be similar to the cross-validated accuracy. The effects
of choosing different values of $k$ are explored in a later section. 

During cross-validation, four different statistics are gathered: the
number of true-positives (TP), false-positives (FP), true-negatives
(TN) and false-negatives (FN). All models were built as binary classifiers,
so that 'positive' refers to membership in cohort 2: the suicide cohort.
Thus, in this case, false-positives are those who were incorrectly
classified as suicidal, whereas false-negatives are patients whose
suicide was not foreseen. These four statistics can be presented in
the form of a two-by-two table, termed the 'confusion matrix'. An
example of such a matrix is shown in table \ref{tab:Example-Confusion-Matrix}.

\begin{table}[h]
\caption{Example Confusion Matrix\label{tab:Example-Confusion-Matrix}}

\begin{centering}
\vphantom{}
\par\end{centering}

\begin{centering}
\begin{tabular}{|c|c|c|}
\cline{2-3} 
\multicolumn{1}{c|}{} & Predicted negatives  & Predicted positives\tabularnewline
\hline 
Expected negatives & Numb. of true negatives (TN) & Numb. of false positives (FP)\tabularnewline
\hline 
Expected positives & Numb. of false negatives (FN) & Numb. of true positives (TP)\tabularnewline
\hline 
\end{tabular}
\par\end{centering}

\vphantom{}

Results are reported in the form of the above confusion matrix. Scores
for this matrix are given by the equations (\ref{eq:accuracy formulas}).

\begin{minipage}[t]{1\columnwidth}%
\rule[0.5ex]{1\columnwidth}{1pt}%
\end{minipage}
\end{table}

In order to be a clinically useful system, it is probably best that,
if the system erred, it did so by finding too many false positives,
rather than by failing to detect a suicidal patient (a false negative).
There are five different result variables that capture this idea in
different ways: the 'recall', 'precision', 'accuracy', $F_{1}$-score
and $F_{2}$-score. The 'recall' addresses the question ``were all
true positives identified (at the risk of some false positives)?''
The 'precision' is the opposite: ``were false positives minimized
(at the risk of failing to identify some true positives)?'' Accuracy,
$F_{1}$and $F_{2}$ are different ways of blending these together
to obtain reasonable composite scores. Presuming that having a high
recall is the clinically desirable way to classify patients, the $F_{2}$-score
is then probably the best quantity to maximize. Note that maximizing
$F_{2}$ can hurt accuracy (\emph{i.e.} too many false positives),
while maximizing accuracy can lead to more false-negatives than might
be desirable.

Mathematically, these five quantities are defined as follows. These
are the standard textbook definitions.

\begin{align}
recall & =\frac{TP}{TP+FN}\nonumber \\
precision & =\frac{TP}{TP+FP}\nonumber \\
accuracy & =\frac{TP+TN}{TP+FP+FN+TN}\label{eq:accuracy formulas}\\
F_{1} & =\frac{2\times precision\times recall}{precision+recall}\nonumber \\
F_{2} & =\frac{5\times precision\times recall}{4\times precision+recall}\nonumber 
\end{align}
Here, TP stands for 'true-positive', as above. All five quantities
can vary between 0 and 1. For a system with a perfect score, all five
quantities would equal 1. If all classification was done by random
chance, then all recall and accuracy would equal 0.5 (for binary classification)
and precision would be the fractional size of the positive group (0.5
if the positive and negative groups are identical in size). As most
of the data analysis concerned groups that were equal in size, it
is desired that all five quantities should be above 0.5. Note that
it is possible to have a classifier that simultaneously scores above
0.5 for some of these measures, and below 0.5 for others.

\subsection*{Ensembles and Voting}

In what follows, the concept of an \emph{ensemble\cite{Opitz1999}}
will be used in two related, but rather distinct ways. In the first
sense, already discussed above, a \emph{model} consists of an ensemble
of \emph{representations}; each representation gets a vote to determine
the final classification that a model makes. In this construction,
the nature of the individual representations can remain rather opaque,
as their effect on the final classification is indirect. In order
to gain more insight into how individual representations combine to
form an ensemble, a restriction is made, in most of what follows,
to limit each model so that it holds only a single representation.
Thus, in most of what follows, the ensemble is overt, and its behavior
is overtly, explicitly presented. That is, the distribution of the
classifications made by each representation, the average behavior,
and the variance, is explicitly presented. Since each model holds
only one representation, the ensemble is referred to as an ensemble
of models.

However, in the end, one wants to revert to the intended purpose of
the ensemble, which is to improve accuracy by combining multiple representations
into one model, and performing classification by majority vote. In
this case, the accuracy of a model will presumably depend on the number
of representations within it. An exploration of how this accuracy
depends on the size of the ensemble is given in the final sections.
To summarize, there are two ensembles: the ensemble of representations
comprising a model, and the ensemble of models.

\subsection*{Ensemble Formalities}

This section sketches a formal, mathematical development of the ensemble
classifier.

Let $P(g|p,m)$ be the probability that a given model $m$ will classify
a given patient $p$ into group $g$. For a fixed patient and model,
this probability is either zero or one (the classifier either assigns
the patient to group $g$ or it doesn't), so that $P(g|p,m)$ is just
the set-membership function:
\[
P(g|p,m)=\mbox{{\bf 1}}_{g|p,m}
\]
The classifier may be incorrect in it's assignment, of course. In
what follows, it is presumed that all classifiers are binary, so that
the group $g$ ranges over the values $\{pos,neg\}$ denoting that
a patient does or does not belong to the group. The law of the excluded
middle is assumed:
\[
P(g=pos|p,m)+P(g=neg|p,m)=1
\]
that is, a given patient is classified as either positive or negative. 

If the patients are divided into a training set and a test set, and
the classifier is trained on the training set, then $P(g|p,m)$ can
be directly measured and evaluated on the test set. Let $S_{pos}$
and $S_{neg}$ be the sets of patients in the test set that are positive
or negative for belonging to the group $g$. Then a given classier
$m$ gives the following counts for true positives (TP), \emph{etc}:
\begin{align*}
TP & =\sum_{p\in S_{pos}}P(g=pos|p,m)\\
FP & =\sum_{p\in S_{neg}}P(g=pos|p,m)\\
FN & =\sum_{p\in S_{pos}}P(g=neg|p,m)\\
TN & =\sum_{p\in S_{neg}}P(g=neg|p,m)
\end{align*}
The above formulas merely provide a more formal definition connecting
two different notations for the same concepts, and nothing more.

The ensemble average is given by
\[
P(g|p)=\frac{1}{\left|M\right|}\sum_{m\in M}P(g|p,m)
\]
where $M$ is the set of models making up the ensemble, and $\left|M\right|$
is the size of this set. In essence, the ensemble average is an expectation
value. Note that the ensemble average is now a real-valued quantity,
ranging over the interval {[}0,1{]}. The \emph{poses} inference command
uses the ensemble average to perform classification, and reports the
average itself as the '\emph{confidence}' of the inference. Specifically,
\[
infer(p)=\begin{cases}
pos & \mbox{if }P(g=pos|p)>0.5\\
neg & \mbox{if }P(g=pos|p)<0.5
\end{cases}
\]
and
\[
confidence(p)=\begin{cases}
2P(g=pos|p)-1 & \mbox{if }P(g=pos|p)>0.5\\
2P(g=neg|p)-1 & \mbox{if }P(g=neg|p)>0.5
\end{cases}
\]
Note that this is closely related to the accuracy (equation \ref{eq:accuracy formulas}
above) on the test set:
\[
accuracy=\frac{1}{\left|S\right|}\left[\sum_{p\in S_{pos}}P(g=pos|p)+\sum_{p\in S_{neg}}P(g=neg|p)\right]
\]
where $\left|S\right|=\left|S_{pos}\right|+\left|S_{neg}\right|$
is the size of the test set.

\section*{Results Overview}

A number of different data analysis experiments were performed. These
include the effect of tuning adjustable parameters on the machine-learning
system, the exploration of ensemble averages, the examination of the
words that appeared in actual models, the effect of data cuts (\emph{i.e.}
excluding infrequent words from the models), and the predictive value
of word-pairs, trigrams and 4-grams.

\subsection*{Example Model}

Running the the classifier once, for a given set of parameters, results
in a single model being created. The precise model, and its accuracy,
depends on the training parameters, such as run-time, the number of
features selected, the number of representations comprising the model,
and other variables. In all cases (for all parameter settings), the
resulting model fits the training data very well. One such case, typical
of all, is shown in table \ref{tab:Training-Confusion-Matrix,}. When
this model is evaluated on the test set, the accuracy and other measures
are, of course, sharply lower. In essence, the model is over-fit on
the train set. The results for the best-fit model on the test set
are shown in table \ref{tab:Test-Confusion-3v2}.

\begin{table}
\caption{Training Confusion Matrix, Group 1 vs. Group 2\label{tab:Training-Confusion-Matrix,}}

\vphantom{}

\begin{centering}
\begin{tabular}{|c|c|c|}
\cline{2-3} 
\multicolumn{1}{c|}{} & Predicted Grp 1 & Predicted Grp 2\tabularnewline
\hline 
Expected Grp 1 & 277 & 3\tabularnewline
\hline 
Expected Grp 2 & 11 & 265\tabularnewline
\hline 
\end{tabular}
\par\end{centering}

\begin{centering}
\vphantom{}
\par\end{centering}

\begin{raggedright}
Confusion matrix, in the form of table \ref{tab:Example-Confusion-Matrix},
for the training set. The model predictions are shown in the columns,
the expected results in rows. There are $4\times(70+69)=556$ training
records to be classified in a 5-fold cross validation. 
\par\end{raggedright}

\begin{centering}
\vphantom{}
\par\end{centering}

\vphantom{}

\begin{centering}
\begin{tabular}{|c|c|c|}
\hline 
Accuracy  &  0.9748 & (542 correct out of 556 total)\tabularnewline
\hline 
Precision & 0.9888 & (265 correct out of 268 total)\tabularnewline
\hline 
Recall & 0.9601 & (265 correct out of 276 total) \tabularnewline
\hline 
FP Rate & 0.0107 & (3 false pos out of 280 total)\tabularnewline
\hline 
F\_1 Score & 0.9743 & \multicolumn{1}{c}{}\tabularnewline
\cline{1-2} 
F\_2 Score & 0.9657 & \multicolumn{1}{c}{}\tabularnewline
\cline{1-2} 
\end{tabular}
\par\end{centering}

\vphantom{}

The results shown here indicate that the model that was created fits
the training data very well, excelling in all measures. This is to
be expected for the training set. The data shown are for a classifier
that distinguishes groups 1 and 2, trained on the bag-of-words dataset.
There were 3000 features pre-selected, 240 features dynamically selected,
and two word-count thresholds used. In practice, these parameters
have almost no effect on the above results: essentially all parameter
settings result in very similar measures.

\begin{minipage}[t]{1\columnwidth}%
\rule[0.5ex]{1\columnwidth}{1pt}%
\end{minipage}
\end{table}

\begin{table}[H]
\caption{Test Confusion Matrix, Group 3 vs. Group 2\label{tab:Test-Confusion-3v2}}

\vphantom{}

\begin{centering}
\begin{tabular}{|c|c|c|}
\cline{2-3} 
\multicolumn{1}{c|}{} & Predicted Grp 1 & Predicted Grp 2\tabularnewline
\hline 
Expected Grp 1 & 43 & 27\tabularnewline
\hline 
Expected Grp 2 & 22 & 47\tabularnewline
\hline 
\end{tabular}
\par\end{centering}

\vphantom{}

Confusion matrix for the test set. The model predictions are shown
in the columns, the expected results in rows. There are $70+69=139$
test records to be classified in a 5-fold cross validation. 

\vphantom{}

\vphantom{}

\begin{centering}
\begin{tabular}{|c|c|c|}
\hline 
Accuracy  & 0.6475 & (90 correct out of 139 total)\tabularnewline
\hline 
Precision & 0.6351 & (47 correct out of 74 total)\tabularnewline
\hline 
Recall & 0.6812 & (47 correct out of 69 total) \tabularnewline
\hline 
FP Rate & 0.3857 & (27 false pos out of 70 total)\tabularnewline
\hline 
F\_1 Score & 0.6573 & \multicolumn{1}{c}{}\tabularnewline
\cline{1-2} 
F\_2 Score & 0.6714 & \multicolumn{1}{c}{}\tabularnewline
\cline{1-2} 
\end{tabular}
\par\end{centering}

\vphantom{}

The results shown here are for a model trained on a set of 3000 pre-selected
features, dynamically narrowed to 500 features during the run. Input
features were created by partitioning the word-counts into 2 levels,
with a threshold at the word-count average. This model was selected
to maximize accuracy, rather than recall or $F_{2}$ score; however,
it appears to have the best $F_{2}$ score of all those explored.

\begin{minipage}[t]{1\columnwidth}%
\rule[0.5ex]{1\columnwidth}{1pt}%
\end{minipage}
\end{table}

Each model consists of ten representations, each resembling that shown
in table \ref{tab:Example-Representation}. Given these representations,
positive and negative keywords can be extracted. Positive keywords
are those that appear in the target group (here, group 2), but not
(as frequently) in the control group. Negative keywords are the reverse:
they appear more frequently in the control group than the target group.
A set of positive keywords, distinguishing groups 2 and 3, are shown
in table \ref{tab:Positive Keywords} and a set of negative keywords
in table \ref{tab:Negative keywords-3v2}.

Note also that a fair number of the keywords apper to be typographic
errors, or are otherwise relatively rare. This can be easily explained:
rare words will appear in relatively few records, and thus their presence
gives an immediate mechanism with which to identify those records.
Unfortunately, this also means that such keywords also have a poor
predictive value: the fact that some word was mis-spelled in a particular
patient record is very unlikely to be of any future use in classifying
other patients. However, it can also be counter-productive to exclude
keywords becuse they do not seem to be obviously relevant. For example,
'ALBUTEROL' sometimes appears among the postive keywords; superficially,
it is an asthma medication, and thus non-predictive and irrelevant.
However, it is also well-known to be associated with suicide risk.
Telling noise from data by examining keywords is not an easy task.
The role of infrequent words is explored in a later section.

\begin{table}[H]
\caption{Positive Keywords, Group 3 vs. Group 2\label{tab:Positive Keywords}}

\vphantom{}

\begin{centering}
\begin{minipage}[t]{0.9\columnwidth}%
AA AAA ABOUT ADULT ANTIPSYCHOTIC APPOINT BARS CAHNGE CANCERS CONSISTANTLY
DESPONDENT DISORDER DRUSEN FRIGHTENING HC ICDCM INTERMITTANT LIPITOR
LUQ MONFRI NALCOHOL NOBODY PRIVATE PUNCTUM REGAINED REORDER RESTRICTED
SHAVE SPARE SPELL STANDARDS STRAIGHTENED STRANGE STREET STVHCS STX
SUBSALICYLATE SWABS TACH TE TELEMETRY TEMAZEPAM THY TIB TP TRANSFUSIONS
TRAVELS TURMOIL TUSCON TWAVE UC ULTIMATELY UNCOOPERATIVE UNDERGONE
UNRESECTABLE URINATED VALLEY VIDEOS VISUALIZATION VTACH WATCHES WHEN
WHIP WORTHLESSNESS WTIH YE YOUNGER ZER%
\end{minipage}
\par\end{centering}

\vphantom{}

The above is a list of positive keywords that distinguish groups 2
and 3. That is, the model requires that these words appear more frequently
in group 2 than group 3. Note the appearance of a fair number of emotionally
laden words. Not all models result in this particular word-list; differences
between different models is discussed in a later section.

\begin{minipage}[t]{1\columnwidth}%
\rule[0.5ex]{1\columnwidth}{1pt}%
\end{minipage}
\end{table}

\begin{table}[H]
\caption{Negative Keywords, Group 3 vs. Group 2\label{tab:Negative keywords-3v2}}

\vphantom{}

\begin{centering}
\begin{minipage}[t]{0.9\columnwidth}%
AFRAID AGGRAVATING ALIGNMENT ALOH ALOT ANT AROUSE BLOCKERS BRONZE
CRYING DEMONSTRATING DENT DIRECTORY DISHEVELED DOCUSATE DPT EFFECTED
EPIDERMAL FEAR GEROPSYCHIATRY MEALTIME NEUT NOCTURIA NOTABLY OBESITY
OUTSTANDING PRACTICING PREFILLED PREOCCUPIED PRESBYOPIA PSYCHIATRICALLY
QUADRANTS RANGES RESIDENTIAL RUMINATES SPECIMEN SPLITTING SSN STAIN
STRUGGLES STUDENT STYLE STYLES SU SUBLUXATION SUPERVISOR SUPERVISORS
SUPPLY SYMPTOMS TEACHER TEASPOONFUL TEETH TEMPOPORMAND TFTS THI TON
TOP TOPICAL TRAZODONE UCINATIONS UES UNCHANGED UNHAPPY UNIQUE UNMARRIED
UNPLEASANT UNSP UNT UPCOMING USEFULNESS VERIFIED VET VIRTUE VISA VISIT
VIT VOIDING VOLUME WALKIN WARNING WARRANT WELLGROOMED WILLING WOUNDED
XPATIENT YEAR %
\end{minipage}
\par\end{centering}

\vphantom{}

The above is a list of negative keywords that distinguish groups 2
and 3. That is, the model requires that these words appear less frequently
in group 2 than group 3. Given that group 3 is the psych patient group,
it is not surprising that many of the words seem to have a psychiatric
significance.

\begin{minipage}[t]{1\columnwidth}%
\rule[0.5ex]{1\columnwidth}{1pt}%
\end{minipage}
\end{table}

\subsection*{Ensemble Averages}

The space of all possible models of a dataset is astronomically large,
and cannot be exhaustively searched. The moses/poses system uses a
pseudo-random number generator to explore different parts of the search
space, both during the genetic algorithm part of the search, and also
during representation construction. The resulting final model thus
depends on the initial random number seed; how well the model scores
will as well. It is not clear how well the score of an individual
model can be trusted, as there is no \emph{a priori} argument that
it will always extend in a good way over a larger dataset. To mitigate
this uncertainty, an ensemble average may be used. In this case, the
average of a large number of models, each built with a different initial
random number seed, may be used. 

\begin{figure}
\caption{Ensemble Average\label{fig:Ensemble-Average}}

\includegraphics[width=0.95\columnwidth]{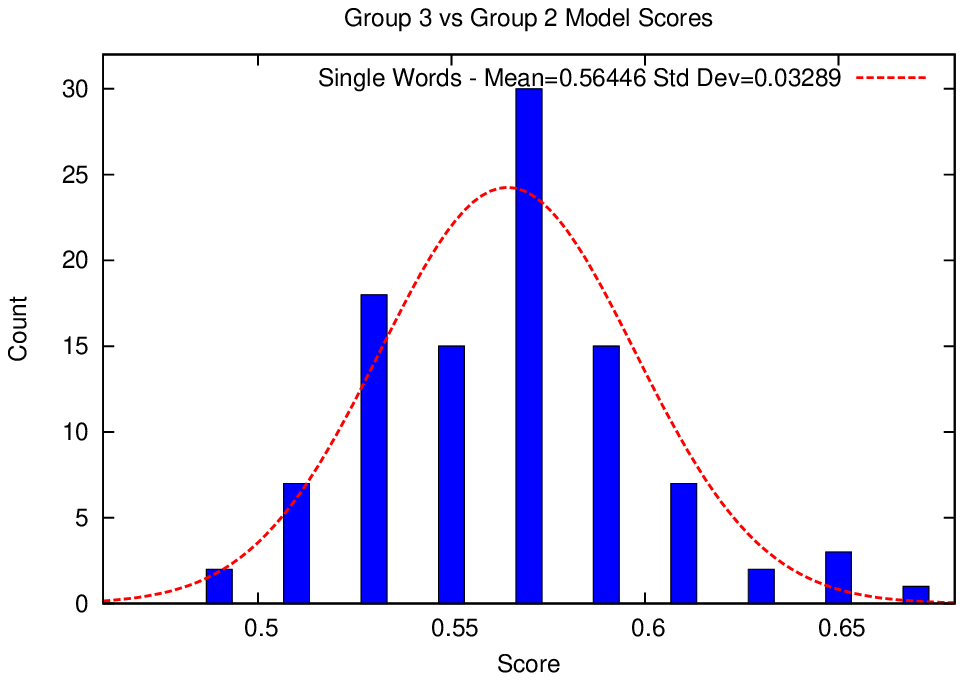}

This bar chart shows the distribution of model accuracy scores for
100 models built to distinguish between the cohort 2 and cohort 3
for the bag-of-words dataset. The accuracy scores of each model was
assigned to a bin that is 0.02 wide; thus there are five bars between
0.5 and 0.6 in this graph. The models were trained with with a single-word
bag-of-words dataset, with word-count thresholding into 4 bins, and
240 dynamically chosen features, out of an initial feature set size
of 3000. The fitted curve is a Gaussian, with a mean of 0.5645 and
a standard deviation of 0.0329. This graph suggests that the 'typical'
accuracy of a single model is then 56.45\%, although there are a few
models that score exceptionally well, including five models with an
accuracy of 64\% or better. It is not clear that the distribution
is in fact Gaussian; it is possible that a log-normal distribution
would provide a better fit. Note that a log-normal distribution would
be centered at the location $\mu=0.5728$.

Note that this same bar chart is shown again in figure \ref{fig:N-grams},
where it is compared to other models. Although this shows the best
bag-of-words model, it is outperformed by all bag-of-phrases models.

\begin{minipage}[t]{1\columnwidth}%
\rule[0.5ex]{1\columnwidth}{1pt}%
\end{minipage}
\end{figure}

In much of what follows, ensemble averages will be used. In all cases,
100 distinct models are built. The figure \ref{fig:Ensemble-Average}
illustrates how this works: it shows a bar-graph of the accuracy scores
of 100 different models created with the same parameters and the same
dataset, differing only in the initial random number seed. The figure
shows a Bell curve fit to this data. A later section looks at model
differences in greater detail.

\subsection*{Cross-Validation}

Classifier performance depends strongly on the choice of the training
set. To obtain an idea of how the training set influences accuracy
scores, several different experiments were performed, summarized in
table \ref{tab:Cross-Validation-Performance}. In all cases, a total
of ten different training/test set partitions were created, by performing
a random draw (that is, patients were chosen randomly to belong to
either the training or the test set). This allows the average accuracy
to be obtained across the ten different test sets, as well as the
standard deviation of the distribution. The data presented is for
a bag-of-word-pairs model, which is presented in greater detail in
later sections.

\begin{table}
\caption{Cross-Validation Performance\label{tab:Cross-Validation-Performance}}

\vphantom{}

\begin{centering}
\begin{tabular}{|c|c|c|}
\cline{2-3} 
\multicolumn{1}{c|}{} & Mean Accuracy & Std. Dev.\tabularnewline
\hline 
80/20 set 1 r0 & 0.635 & 0.098\tabularnewline
\hline 
80/20 set 1 r1 & 0.573 & 0.075\tabularnewline
\hline 
80/20 set 2  & 0.612 & 0.081\tabularnewline
\hline 
75/25 & 0.560 & 0.114\tabularnewline
\hline 
66/33 & 0.573 & 0.103\tabularnewline
\hline 
50/50 & 0.559 & 0.0510\tabularnewline
\hline 
\end{tabular}
\par\end{centering}

\vphantom{}

Average accuracy and standard deviation over ten different training/test
set partitions. The 80/20 partition allocates 80\% of the patients
to the training set, and 20\% to the test set. Three different experiments
are shown: ``set 1 r0'' and ``set 1 r1'' both use exactly the
same partitions, but initialize the learner with two different random
seeds. The ``set 2'' experiment explores accuracy over a a completely
different set of ten partitions. The rows labeled 75/25, 66/33 and
50/50 explore the effects of reducing the size of the training set. 

\begin{minipage}[t]{1\columnwidth}%
\rule[0.5ex]{1\columnwidth}{1pt}%
\end{minipage}
\end{table}

The overall suggestion from table \ref{tab:Cross-Validation-Performance}is
that maximizing the size of the training set, and then making up for
the small size of the test set by averaging over many partitions,
is the best strategy. For the remainder of the analysis, an 80/20
split, averaged over five round-robin partitions, will be used: this
is the 5-fold cross validation.

\subsection*{Parameter Tuning}

One of the most time-consuming experiments is to determine the optimal
settings for the training parameters. The two most important and sensitive
of these are the number of bins chosen for word-counts, and the number
of dynamic features. The size of the static feature list seems to
have little bearing on the ultimate score, once this is reasonably
large; a static feature set of 3000 seems to be sufficient. Nor does
the total training time seem to matter much, once it is sufficiently
long. Increasing the training time will cause the system to build
ever-more complex models, attempting to attain a perfect score on
the training set. These more complex models do not appear to score
better on the test set, nor do they appear to score any worse, either.

One noteworthy effect, though, is that the larger the dataset size,
the less sensitive the results are on these adjustable parameters.
Roughly speaking, the parameters are used to 'focus' on the distinctive
parts of the dataset, in much the same way that image processing is
used to sharpen an image. For the larger datasets, there seems to
be less of a need to 'focus'; but whether this is a real effect or
an artifact is unclear. All of the datasets are small, and the largest
dataset is about three times the size of the smallest one.

\begin{figure}
\caption{Parameter Variation\label{fig:Parameter-Variation}}

\includegraphics[width=0.95\columnwidth]{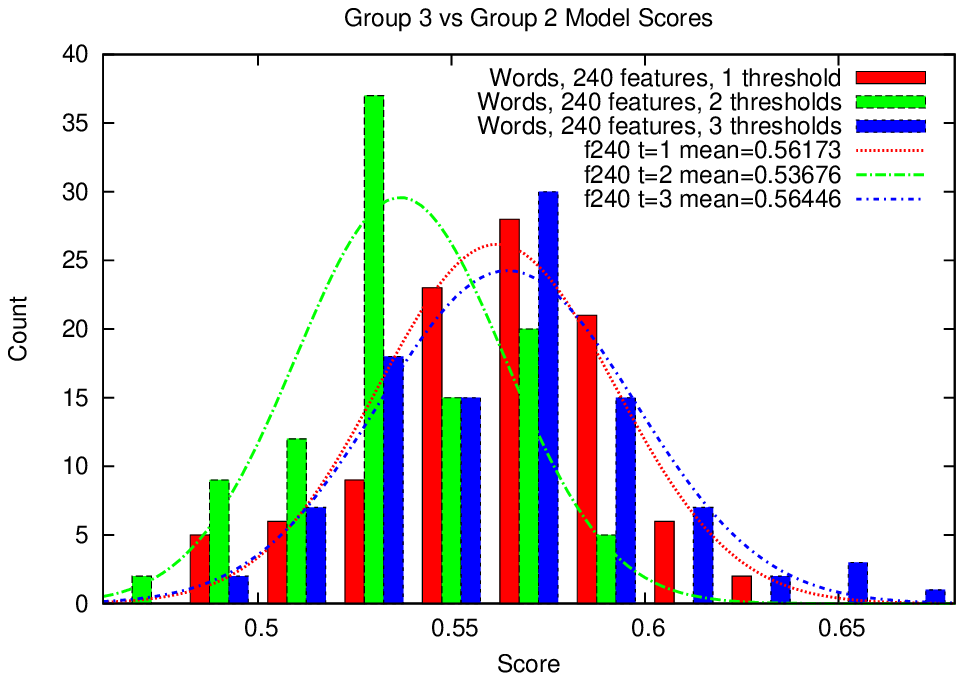}

An example of accuracy score distributions for three different parameter
settings. All three bar-graphs are built from the same dataset, the
bag-of-words dataset that distinguishes groups 2 and 3. All three
sets of models were trained so that 3000 features are pre-selected,
and 240 features are dynamically selected during run-time. The only
difference is whether the word-counts were binned into 2, 3 or 4 bins
by 1, 2 or 3 thresholds. When one threshold is used, it is always
set at the mean word count. When two thresholds are used, they are
set one standard deviation above and below the mean word count. When
three thresholds are uses, they are set at the mean, and one standard
deviation above and below. Observe that almost all of the classification
effect is derived from using just one threshold: adding two more improves
classification, but only slightly. Observe that the threshold located
at the mean appears to be the most important; when it is not used,
classification suffers. This is not always the case; for some of the
parameter settings, such as those where more or fewer dynamic-runtime
features are used, the situation is reversed: an even number of thresholds
work better than an odd number. Nor is it the case that adding more
thresholds always improves the score; sometimes, this leads to over-training
instead, as is evident in table \ref{tab:Tuning-the-classifier,}.

\begin{minipage}[t]{1\columnwidth}%
\rule[0.5ex]{1\columnwidth}{1pt}%
\end{minipage}
\end{figure}

In order to evaluate the effect of this parameter tuning, ensemble
averages, over 100 models, were used, as described above. For each
set of parameters, the mean and standard deviation of the accuracy
distribution was computed. These, as a function of the parameters,
are shown in table \ref{tab:Tuning-the-classifier,}. The figure \ref{fig:Parameter-Variation}
shows three typical distributions from this table, one of which was
already shown in figure \ref{fig:Ensemble-Average}.

\begin{table}
\caption{Tuning the classifier, Group 3 vs. Group 2\label{tab:Tuning-the-classifier,}}

\begin{centering}
\vphantom{}
\par\end{centering}

\begin{centering}
\begin{tabular}{|c||c|c||c|c||c|c|}
\hline 
\multirow{2}{*}{Feat} & \multicolumn{2}{c||}{Num Thresh=1} & \multicolumn{2}{c||}{Num Thresh=2} & \multicolumn{2}{c|}{Num Thresh=3}\tabularnewline
\cline{2-7} 
 & Mean Acc & Std Dev & Mean Acc & Std Dev & Mean Acc & Std Dev\tabularnewline
\hline 
\hline 
40 & 0.5688 & 0.0234 &  &  &  & \tabularnewline
\hline 
50 & 0.5803 & 0.0243 &  &  &  & \tabularnewline
\hline 
60 & 0.5901 & 0.0261 &  &  &  & \tabularnewline
\hline 
75 & 0.5942 & 0.0268 &  &  & 0.5699 & 0.0256\tabularnewline
\hline 
90 & 0.5879 & 0.0366 &  &  & 0.5359 & 0.0302\tabularnewline
\hline 
120 & 0.5898 & 0.0265 &  &  & 0.5698 & 0.0265\tabularnewline
\hline 
150 & 0.5717 & 0.0259 &  &  & 0.5761 & 0.0272\tabularnewline
\hline 
180 & 0.5823 & 0.0276 & 0.5531 & 0.0219 & 0.5841 & 0.0322\tabularnewline
\hline 
240 & 0.5617 & 0.0305 & 0.5368 & 0.0270 & 0.5645 & 0.0334\tabularnewline
\hline 
360 & 0.5629 & 0.0334 & 0.5178 & 0.0276 & 0.5496 & 0.0290\tabularnewline
\hline 
500 & 0.5309 & 0.0313 &  &  & 0.5116 & 0.0290\tabularnewline
\hline 
\end{tabular}
\par\end{centering}

\begin{centering}
\vphantom{}
\par\end{centering}

This table shows the effect, on the mean accuracy, of tuning the classifier
parameters. All entries in the table are for models built from the
same dataset, the bag-of-words dataset that distinguishes groups 2
and 3. The table shows the mean accuracy and standard deviation for
the 5-fold validation of 100 models. All models were trained so that
3000 features are pre-selected. The number of features dynamically
selected during run-time are indicated in the first column. The thresholds
are used to bin word-counts into 2, 3 or 4 bins by 1, 2 or 3 thresholds.
When one threshold is used, it is always set at the mean word count.
When two thresholds are used, they are set one standard deviation
above and below the mean word count. When three thresholds are uses,
they are set at the mean, and one standard deviation above and below.
Histograms for the 240-feature case are shown in figure \ref{fig:Parameter-Variation}.
It seems that using only one threshold is usually, but not always
the best. The dependence on the number of dynamical features is somewhat
uneven. The first three columns are graphed below.\\
\\

\centering{}\includegraphics[width=0.95\columnwidth]{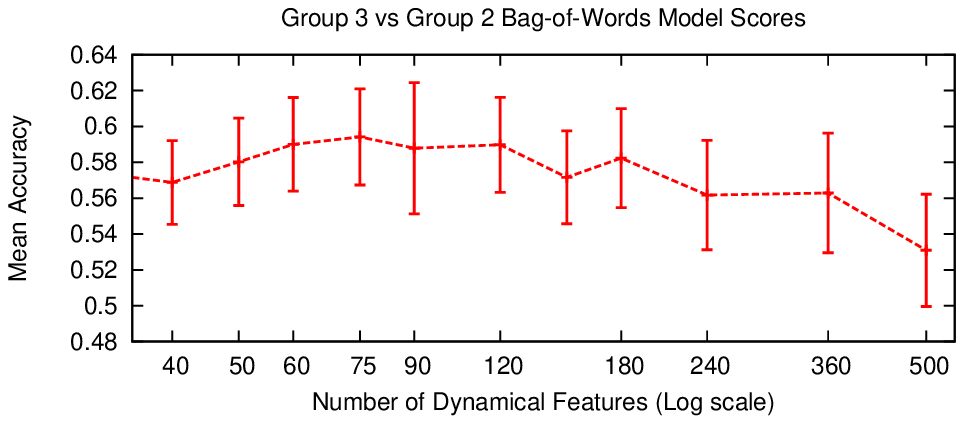}\\
\end{table}

\subsection*{Infrequent Words}

Infrequently occurring words appear to play an important predictive
role. By 'infrequent' it is meant words that appear in less than a
fourth of the patient records, and possibly in as few as just two.
This is quite a remarkable result, and it manifests itself in several
ways in the data. It raises questions: is this an artifact of working
with sparse data, or is it possible that suicidal patients present
in a variety of ways, with no common set of symptoms? This section
explores how infrequent words influence model construction and model
accuracy.

When using ensemble averages, the question arises: how similar or
different are the individual models? Do they differ only a little,
or a lot? Measuring precise differences is difficult, due to the fact
that the actual representations (table \ref{tab:Example-Representation})
are Boolean program trees. However, a general indication can be had
by generating keyword lists, and comparing them between models.

\begin{figure}
\caption{Common Keywords\label{fig:Common-Keywords}}

\begin{centering}
\vphantom{}
\par\end{centering}

\includegraphics[width=0.95\columnwidth]{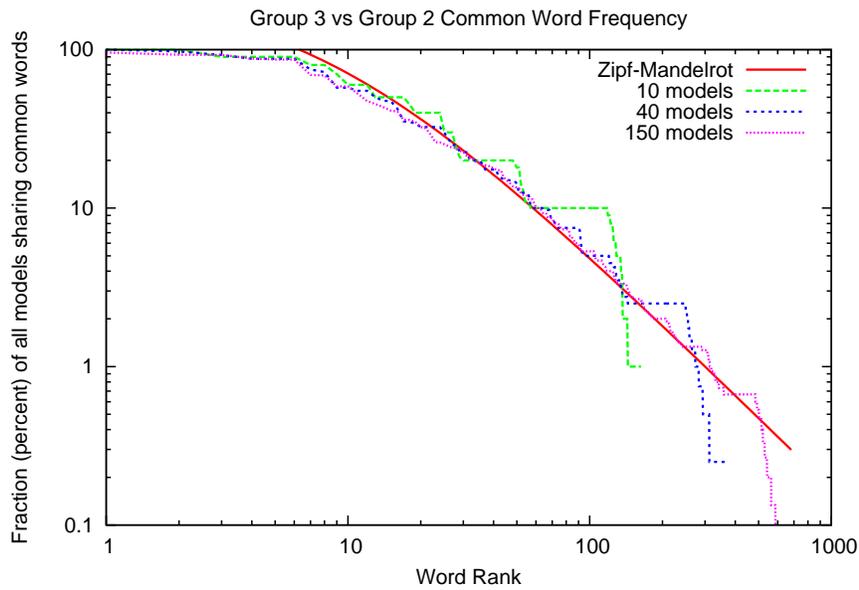}

\begin{centering}
\vphantom{}
\par\end{centering}

This graph shows the fraction of all representations that share words
in common. Thus, the highest ranked word is used in the largest number
of representations, the next highest ranked word is used in the next
greatest number of representations, and so on. For the set of 10 models,
there were 100 representations, which used 163 unique words among
them, sharing many of them. The set of 40 models (400 reps) used 371
unique words, while the set of 150 models (1500 reps) used 682 unique
words. Thus, creating more models does cause more words to be employed,
but at a diminishing rate. The smooth line labeled 'Zipf Mandelbrot'
is a rough fit to the data, given by the formula $\sim(\mbox{rank}+8)^{-1.5}$
. The result here is phenomenological.

\begin{minipage}[t]{1\columnwidth}%
\rule[0.5ex]{1\columnwidth}{1pt}%
\end{minipage}
\end{figure}

This was done by generating N different models, with each model consisting
of 10 representations. Each model is generated by using exactly the
same parameters, but differing only in the initial random number seed:
thus, these are true ensembles. One then asks: how many words are
shared by all representations? How many are shared by most representations?
How many are shared by only half of them? The answer to these questions
is that two words are shared by all representations, six are shared
by 90\% of all representations, and 17 by more than half of all representations.
Rather surprisingly, these counts do not depend much on the number
N of different models: one can look at 10 or 150 models, and the number
of words shared in common stays more or less the same. This is shown
in figure \ref{fig:Common-Keywords}, where the percentages are graphed
for N=10, 40 and 150 models (100, 400 and 1500 representations).

If may be the case that many of these words are acting as if they
were synonyms for one another: not in the literal sense of having
the same meaning, but rather that they are being used in similar contexts.
Perhaps there is a common set of words that are indicative, but some
patient records only use some of them, while others use others. But
perhaps, there is a different situation: when a patient record has
one of these words, it also has many of the others. In such a case,
it would be enough to pick just one of these words out to build a
model, and if different models picked different words, its is only
because they are inter-changeable, and the models are only superficially,
but not deeply, different. A cluster analysis would need to be performed
to determine this. 

\begin{figure}
\caption{Model Words Rank\label{fig:Model-Words-Rank}}

\includegraphics[width=0.95\columnwidth]{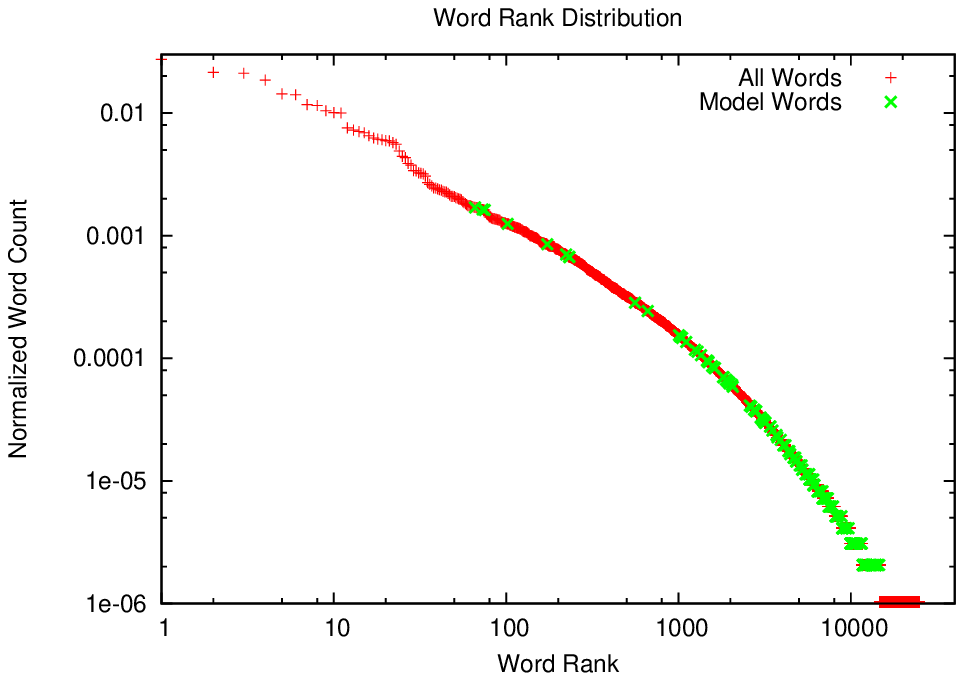}

This graph reproduces that of figure \ref{fig:Word-Rank-Distribution},
high-lighting the words that were used to construct models in green.
In total, there are 163 words highlighted in green, taken from the
N=10 model collection. It is clear that, with a few exceptions, most
of the words used to distinguish patient groups are words that are
infrequently used. The red bar in the lower right corresponds to words
that appear only once amongst all the patient texts. The next green
bar above that corresponds to words that appear twice, and so on.
Thus, this graph makes clear that words that appear only a small number
of times (2 through 30) are all vital for distinguishing patient groups.
Note that, although the green crosses appear to dominate the lower
right of the graph, this is partly an illusion: there are 163 green
crosses in total, whereas there are more than ten thousands red crosses
to the lower right. Thus, although words that appear only twice in
the text are vital for model building, only a tiny fraction of these
are actually used.

\begin{minipage}[t]{1\columnwidth}%
\rule[0.5ex]{1\columnwidth}{1pt}%
\end{minipage}
\end{figure}

Out of the collection of all words, what is the rank of the words
chosen for use in a model? This is hinted at in figure \ref{fig:Model-Words-Rank}.
As can be clearly seen, infrequently-used words are vital for distinguishing
patient groups. Indeed, it would appear that distinguishing words
all have fairly small counts (2 through 30 or 40), with a few exceptions.
Observe that not all rare words are used for model building: there
are tens of thousands of words that appear less than 5 times in the
text; of these, less than a few hundred are selected for use in a
model.

However, this dependence on rare words for model building indicates
that the system is keying on attributes that are shared by only handfuls
of patients. It is not clear if this is an artifact of the small dataset
size, or whether different patients are showing distinct, non-overlapping
'symptoms'. Recall that there are only 70+69=139 patients in total
that are being discriminated between by these models. Thus, if a word
appears only 10 times in total in the entire text, then this word
can select at most only 10 patients (unless it is a negative keyword,
in which case it can be used to rule out 139-10=129 patients). Is
this happening because there are 10 patients who are presenting in
a very specific way? Or is this because the records are sparse, and
that perhaps all patients would present in this way, but it was simply
not observed and noted? In other words, do all suicidal patients present
in the same way, or are there classes of distinct behavior patterns?
If there is commonality to all suicidal behavior, it is not particular
evident in this data.

A different measure of the importance of infrequent words can be obtained
by excluding them from model building: that is, by creating models
from word lists that include only those words that occur 2 or 4 or
more times in the text. Superficially, this seems like a wise idea.
If a word appears in only one patient record, and it is found during
the training phase, then it is impossible that this word will also
appear in one of the test patient records. Thus, it cannot contribute
to the accuracy of the model on the test set: both the positive and
negative cohorts will be missing this word; it has no predictive value.
If a word appears in only two patient records, then it is unlikely
that one of those locations will be in the test-set (due to the 4/5'ths
- 1/5'th split). Thus, one might also believe that such words have
little or no predictive value. Perhaps accuracy can be increased by
cutting down the dataset, and discarding all words that appear fewer
than M times in the dataset. But this is very much not the case. Results
are shown in table \ref{tab:Data-Cuts}.

\begin{table}
\caption{Data Cuts\label{tab:Data-Cuts}}

\begin{centering}
\vphantom{}
\par\end{centering}

\begin{centering}
\begin{tabular}{|c|c|c|c|c|}
\hline 
\multirow{2}{*}{Cut} & \multicolumn{2}{c|}{thresh=1} & \multicolumn{2}{c}{thresh=3}\tabularnewline
\cline{2-5} 
 & Mean & Std. Dev. & Mean & Std. Dev.\tabularnewline
\hline 
\noalign{\vskip3pt}
\hline 
none & 0.5571 & 0.0297 & 0.5645  & 0.0329\tabularnewline
\hline 
2 & 0.5590 & 0.0298 & 0.5578 & 0.0270\tabularnewline
\hline 
4 & 0.5406 & 0.0318 & 0.5343 & 0.0296\tabularnewline
\hline 
8 & 0.5145 & 0.0331 & 0.5153 & 0.0279\tabularnewline
\hline 
16 & 0.4996 & 0.0313 & 0.5109 & 0.0288\tabularnewline
\hline 
\end{tabular}
\par\end{centering}

\begin{centering}
\vphantom{}
\par\end{centering}

This table shows ensemble averages for the accuracy, when infrequent
words are cut from the dataset. Thus, the row labeled '4' indicates
results when all words appearing 4 or fewer times have been cut from
the dataset. All results are, as usual, for an ensemble of 100 models.
All models were trained on the same set of parameters: 3000 statically
selected features, 240 dynamically selected features, and 1 or 3 thresholds,
as indicated. This is the parameter choice that results in the highest
score when no cuts are made, as shown in figure \ref{fig:Ensemble-Average},
figure \ref{fig:Parameter-Variation} and table \ref{tab:Tuning-the-classifier,}.
Cutting words that appear only once is the same as cutting none at
all.

\begin{minipage}[t]{1\columnwidth}%
\rule[0.5ex]{1\columnwidth}{1pt}%
\end{minipage}
\end{table}

Cutting rare words decreases model accuracy. A modest cut of even
4 words has a large impact on scores, and cutting more than that essentially
wipes out the predictive accuracy of a model almost completely.

\subsection*{Word Pairs and N-grams}

A common issue that arises when machine learning is applied to sentiment
analysis is that positive and negative keywords can be negated in
the text, inverting their meaning. So, for example, a keyword ``unhappy''
may occur in a sentence ``he is not unhappy.'' Another issue is
that semantic meaning is not confined to single words, but can be
associated with word pairs, collocations (set phrases) and idioms.
Looking at merely one word of a collocation may imply a less refined
meaning (``intramuscularly'' is not specific the way that ``left
deltoid intramuscularly'' is) or possibly a completely different
meaning altogether (``disturbances'' has psychological overtones;
``visual disturbances'' suggest something completely different:
glaucoma or macular degeneration. Similarly: ``alcohol'' vs. ``alcohol
screening test'', ``pain'' vs. ``pain relieving'', \emph{etc}.)
Thus, one might expect greater predictive value arising from using
neighboring word pairs or even perhaps entire phrases\cite{Wallach2006}.
This is indeed the case, as demonstrated in this section.

In order for this technique to work, one must be careful to apply
appropriate cuts to the dataset. Simply including all possible word
pairs does not improve model accuracy. The reason for this is well-known:
by including word-pairs, the number of candidate features that might
fit the data enlarges to a much larger number. Statistical chance
means that some of these may correlate strongly with the training
set, even though they are not actually predictive. Discarding word
pairs with a low mutual information (MI) score is an obvious cut to
make; one may also contemplate discarding infrequent word pairs, although
experience with single words suggests that this is not a good idea.

An alternative to discarding word pairs with low MI is to consider
only those word pairs that involve a word that has previously been
identified as being 'predictively significant', that is, a word that
already occurs in a single-word model, such as those in tables \ref{tab:Positive Keywords}
and \ref{tab:Negative keywords-3v2}. Word pairs constructed from
these words are 'clinically interesting', in that they provide a larger
window into the notes occurring in a patient record. In this sense,
the approach is inspired by the central idea of corpus linguistics:
in order to better understand the meaning of a word, it is best to
view it in context, to see how it is being used. Inspired by this
idea, it is reasonable to contemplate using three-word phrases (trigrams)
and 4-word phrases (4-grams) to construct a bag-of-phrases. In what
follows, these n-grams will be referred to as 'corpus n-grams', indicating
that they were constructed from 'clinically interesting' words. This
is to draw a distinction between these, and the set of all n-grams
cut down by MI scores. This sort of an approach is known to provide
a positive benefit for classification\cite{Goertzel2006}. 

To create the list of 'significant words', an ensemble of 40 models
were trained on the group 3 vs. group 2 dataset. As noted previously,
in the caption to figure \ref{fig:Common-Keywords}, this ensemble
results in 371 unique words. The set of corpus n-grams were then selected
by considering only those n-grams that contained one of these 371
words.

In what follows, n-grams are constructed not only from n adjacent
words, but also from adjacent words with 'holes' (wild-cards) in them.
The reason for doing this is to properly take into account multi-word
noun and verb modifiers. Thus, for example, the phrase ``horrible
frightening experience'' is composed to two semantically interesting
units: ``horrible\_experience'' and ``frightening\_experience''
(this example is taken from the actual dataset) . The first would
not be captured if one limited oneself solely to adjacent words when
creating pairs. Likewise, when constructing 3-grams, not only were
three adjacent words considered, but also all possibilities for picking
three words out of a string of four consecutive words. When creating
4-grams, all possibilities for picking 4 words out of 5 consecutive
words were considered. When a bag-of-n-grams is constructed, it also
includes those n-grams that are shorter: thus, the bag-of-pairs also
includes single words, and the bag-of-trigrams also includes pairs
and single words in it. Thus, during model building, a pair or a trigram
is used only if it results in a better model than using some individual
word. 

This idea of allowing holes in n-gram construction is to partly emulate
the action of a syntactic parser, which would be able to identify
meaningful semantic relationships, such as adjective-noun or even
subject-verb. In place of syntactic parsing, high mutual-information
phrases can help identify meaningful phrases, and in some ways, can
even be superior, given the fractured, badly-structured and non-grammatical
content of the notes. Unfortunately, even this approach is insufficient
to deal with long-range correlations between words in the text. The
example given above occurs in a note as a part of a semi-automated
system for PTSD screening, whose full content, as it appears in the
note, is: ``\emph{Have you ever had any experience that was so frightening,
horrible, or upsetting that, IN THE PAST MONTH you: Have you had any
nightmares about it or thought about it when you did not want to?
NO}''. This note appears for a control-group patient; the presence
of such semi-automatically generated notes adds to the classification
challenge. It is not clear how to extract this kind of information;
the challenge is similar in some ways to that of anaphora resolution
(the word '\emph{NO}' being the resolution to the question) , and
perhaps techniques from that area could be applied.

\begin{figure}
\caption{N-grams\label{fig:N-grams}}

\includegraphics[width=0.95\columnwidth]{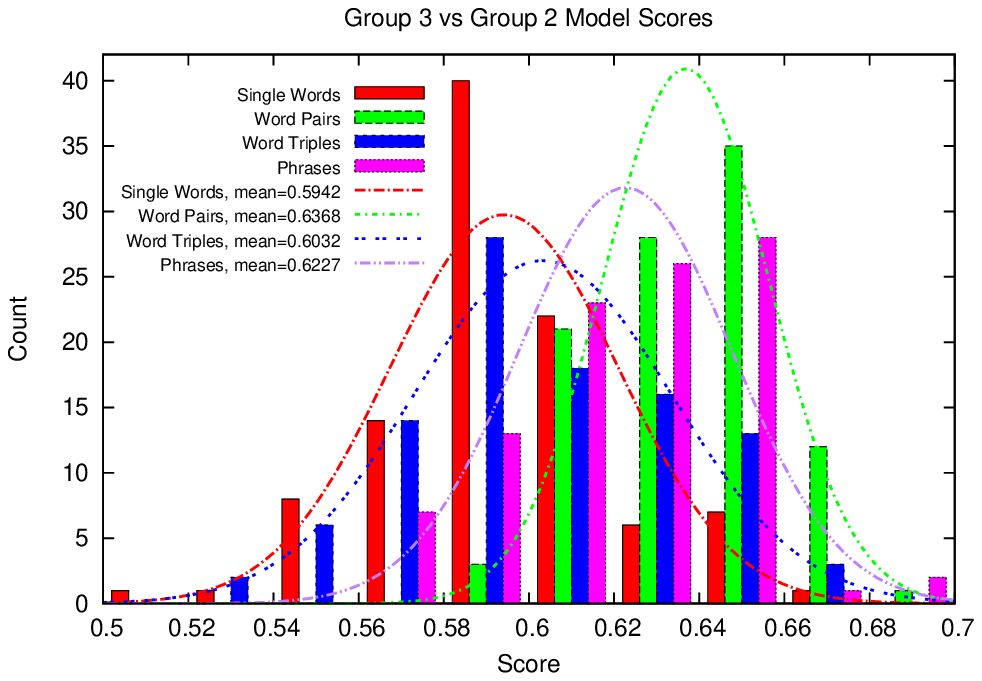}

This bar chart compares three different bag-of-phrases models to the
highest performing bag-of-words model. Observe that all of the bag-of-phrases
models outperform the best bag-of-words model. Results of parameter
tuning are shown in table \ref{tab:Bag-of-Phrases}.

\begin{minipage}[t]{1\columnwidth}%
\rule[0.5ex]{1\columnwidth}{1pt}%
\end{minipage}
\end{figure}

The corpus-linguistics-inspired approach, of considering only those
word phrases that contain words that were previously identified as
'significant', works very well. This is illustrated in figure \ref{fig:N-grams},
where the results of four different experiments are shown: the best
bag-of-words result, the best corpus-pairs result, the best corpus-trigram
result and the best corpus 4-gram result. Particularly noteworthy
is that all of the bag-of-phrases models perform better than the best
bag-of-words model. Of all these, the most outstanding are the word-pairs
results. 

\begin{table}
\begin{centering}
\caption{Bag of Phrases Tuning Results\label{tab:Bag-of-Phrases}}
\vphantom{}
\par\end{centering}

\begin{centering}
\begin{tabular}{|c|c|c|c|c|}
\hline 
Num. Dyn. Feat & single words & word pairs & trigrams & 4-grams\tabularnewline
\hline 
\hline 
2 & 0.5301 & 0.5794 & 0.5642 & 0.5537\tabularnewline
\hline 
3 & 0.5112 & 0.5983 & 0.5473 & 0.5589\tabularnewline
\hline 
5 & 0.5168 & 0.6283 & 0.5649 & 0.5635\tabularnewline
\hline 
8 & 0.5305 & 0.6069 & 0.5440 & 0.5442\tabularnewline
\hline 
12 & 0.5065 & 0.5981 & 0.5551 & 0.5947\tabularnewline
\hline 
16 & 0.5450 & 0.5823 & 0.5614 & 0.6227\tabularnewline
\hline 
24 & 0.5284 & 0.5691 & 0.5615 & 0.5532\tabularnewline
\hline 
32 & 0.5735 & 0.5917 & 0.5546 & 0.6020\tabularnewline
\hline 
40 & 0.5688 &  &  & \tabularnewline
\hline 
50 & 0.5803 & 0.5914 & 0.6032 & 0.5981\tabularnewline
\hline 
60 & 0.5901 & 0.5868 &  & \tabularnewline
\hline 
75 & 0.5942 & 0.5899 & 0.5887 & 0.5801\tabularnewline
\hline 
90 & 0.5879 & 0.6227 & 0.5721 & 0.6019\tabularnewline
\hline 
105 &  & 0.5769 &  & \tabularnewline
\hline 
120 & 0.5898 & 0.5772 & 0.5817 & 0.5706\tabularnewline
\hline 
150 & 0.5717 & 0.5884 &  & \tabularnewline
\hline 
180 & 0.5823 & 0.5906 & 0.5727 & 0.5792\tabularnewline
\hline 
240 & 0.5617 & 0.6036 & 0.5746 & 0.5773\tabularnewline
\hline 
300 &  & 0.6028 &  & \tabularnewline
\hline 
360 & 0.5629 &  & 0.5891 & 0.5784\tabularnewline
\hline 
400 &  & 0.5911 &  & \tabularnewline
\hline 
500 & 0.5309 & 0.5715 & 0.5816 & 0.5655\tabularnewline
\hline 
\end{tabular}
\par\end{centering}

\begin{centering}
\vphantom{}
\par\end{centering}

\begin{centering}
This table shows ensemble averages for the accuracy as the number
of run-time dynamic features was varied. As usual, the ensemble consists
of 100 models, made from 3000 pre-selected features; only one threshold
is used. Single-word results are identical to those reported in table
\ref{tab:Tuning-the-classifier,}. The best ensembles from each column
are graphed in figure \ref{fig:N-grams}. The datasets were subjected
to cuts: All words and n-grams that occurred 4 or fewer times were
cut, as were all n-grams with an MI of less than 2. The chart below
graphs the table contents. Bars showing the standard deviation are
plotted only for the single-word scores; those for phrases are comparable
or smaller. \\

\par\end{centering}

\begin{centering}
\includegraphics[width=0.95\columnwidth]{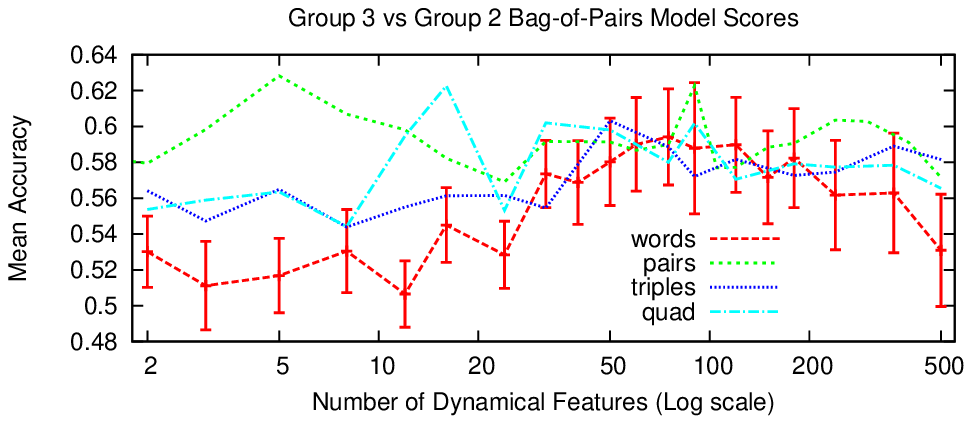}\\

\par\end{centering}

\end{table}

Creating the set of corpus-pairs requires having previously computed
a list of 'significant words'. Creating that list is time-consuming,
since it requires training an ensemble, extracting the words, and
then training again, with pairs. Thus, it is natural to ask if there
are simpler ways of obtaining a list of 'significant words' that are
just as good. There are: in fact, simple single-word feature selection
is sufficient to create a list of 'significant words' that is every
bit as good as that obtained from the ensemble, and maybe even a little
bit better, as shown in table \ref{tab:Choosing-Significant-Words}.
To make the two methods comparable, a simple mutual-information-maximizing
feature selection step was performed to select 371 words, the same
number of words as obtained from the ensemble. Feature selection runs
in seconds, whereas training an ensemble of 40 models can take hours.

\begin{table}
\caption{Choosing Significant Words\label{tab:Choosing-Significant-Words}}

\begin{centering}
\vphantom{}
\par\end{centering}

\begin{centering}
\begin{tabular}{|c|c|c|}
\hline 
Word Source & Mean Accuracy & Std. Dev.\tabularnewline
\hline 
\hline 
Ensemble & 0.5986 & 0.0324\tabularnewline
\hline 
Feature Selection & 0.6036 & 0.0295\tabularnewline
\hline 
\end{tabular}
\par\end{centering}

\begin{centering}
\vphantom{}
\par\end{centering}

A comparison of word-pair ensemble accuracy results for sets of corpus
word-pairs created from two different sources of 'significant words'.
The ensemble 'significant words' consist of 371 words taken from an
ensemble of 40 models. The feature-selection 'significant words' consist
of 371 words selected by maximizing mutual information between the
word counts and the cohort id. The ensembles were trained with the
same parameters as reported above: 3000 static features, 240 dynamic
features, 1 threshold, on the group 3 vs. group 2 dataset.

\begin{minipage}[t]{1\columnwidth}%
\rule[0.5ex]{1\columnwidth}{1pt}%
\end{minipage}
\end{table}

\begin{figure}
\caption{Distribution of Mutual Information\label{fig:Distribution-of-Mutual}}

\includegraphics[width=0.95\columnwidth]{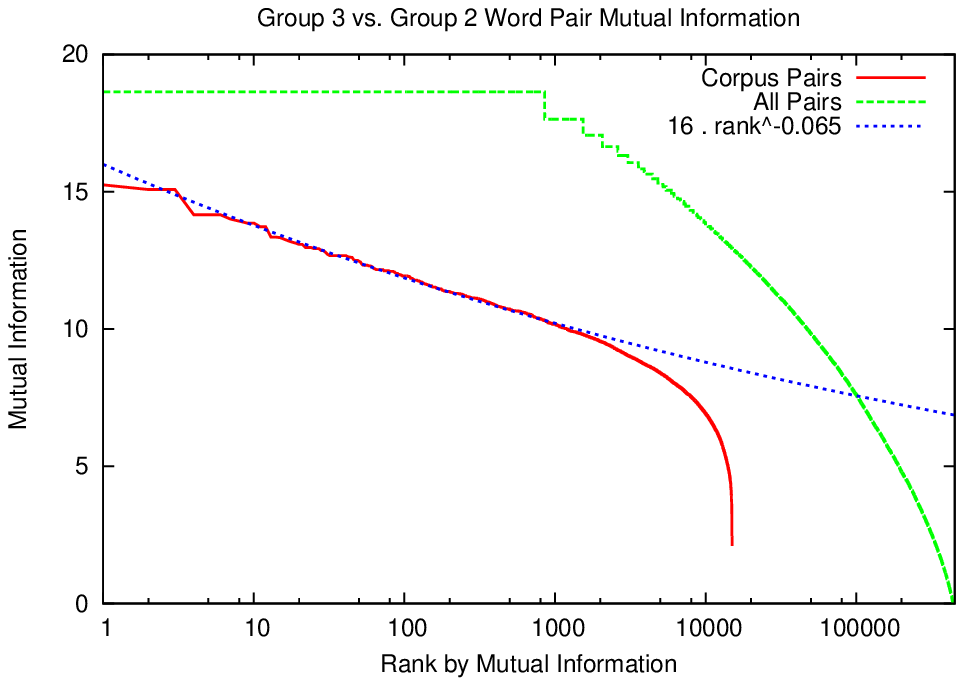}\\

The distribution of corpus-pairs and all word pairs, ranked in decreasing
order of mutual information. The distributions are rather dramatically
different; the corpus-pairs distribution having a Zipfian segment
which is lacking in the all-pairs distribution. The relative rank
of a pair can be obtained by drawing a horizontal line across the
two curves: the corpus-pairs set eliminated all of the high-mutual-information
pairs, as well as most of the low-mutual information pairs.

\begin{minipage}[t]{1\columnwidth}%
\rule[0.5ex]{1\columnwidth}{1pt}%
\end{minipage}
\end{figure}

The distribution of corpus-pairs vs. all-pairs is dramatically different,
as shown in figure \ref{fig:Distribution-of-Mutual}. By selecting
corpus pairs, thousands of the highest-MI pairs are discarded, as
well as most of the low-MI pairs, as well. Perhaps it is possible
to replicate the corpus-pairs results by applying a simple cut to
the all-pairs dataset, and merely discarding the low-MI pairs? This
does not seem to be the case, as shown in table \ref{tab:All-Word-Pair-Cuts}.
This table compares a bag-of-words model to several different all-pairs
models, with different MI cuts applied. Including all pairs does not
improve the score over the bag-of-words. By cutting out low-MI pairs,
the score can be improved somewhat, but the effect is not dramatic;
certainly not as strong as the decision to use corpus-pairs.

\begin{table}
\caption{All-Word-Pair Cuts\label{tab:All-Word-Pair-Cuts}}

\begin{centering}
\vphantom{}
\par\end{centering}

\begin{centering}
\begin{tabular}{|c|c|c|}
\cline{2-3} 
\multicolumn{1}{c|}{} & Mean Accuracy & Std. Dev.\tabularnewline
\hline 
Single Words & 0.5617 & 0.0305\tabularnewline
\hline 
All Pairs & 0.5602 & 0.0418\tabularnewline
\hline 
All Pairs MI>0 & 0.5764 & 0.0275\tabularnewline
\hline 
All Pairs MI>2 & 0.5646 & 0.0282\tabularnewline
\hline 
All Pairs MI>4 & 0.5751 & 0.0277\tabularnewline
\hline 
All Pairs MI>6 & 0.5655 & 0.0281\tabularnewline
\hline 
\end{tabular}
\par\end{centering}

\begin{centering}
\vphantom{}
\par\end{centering}

Ensemble averages for accuracy, comparing different cuts to the bag-of-single-words
accuracy. Simply including all word pairs does not improve the quality
of the model. Discarding those pairs with low mutual information scores
does, however, have a net positive effect. The best location for the
MI cut is not entirely clear, though. All ensembles were trained with
3000 pre-selected features, 240 dynamically selected features, and
one count threshold. (Note that these training parameters do not result
in either the best-words or the best-pairs models, and so should not
be directly compared to those results).

\begin{minipage}[t]{1\columnwidth}%
\rule[0.5ex]{1\columnwidth}{1pt}%
\end{minipage}
\end{table}

The original bag-of-phrases results shown in tables \ref{tab:Bag-of-Phrases},
\ref{tab:Choosing-Significant-Words} and figure \ref{fig:N-grams}
were made by employing some arbitrary, 'intuitive' cuts for the number
of words, and for mutual information. Later experiments on the effect
of cutting rare words shows a net negative effect, as documented in
table \ref{tab:Data-Cuts}. Perhaps it is a mistake, then to cut rare
words and rare word-pairs, when using a bag-of-pairs model? It doesn't
seem so: table \ref{tab:Mutual-Information-Cuts} shows a counter-intuitive
result. In this experiment, no rare words or pairs were cut; only
the cut for the MI was altered. None of the results approach the best
accuracy from table \ref{tab:Choosing-Significant-Words}. Thus, somehow,
when word-pairs come into play, failing to cut rare words and phrases
makes things worse!

\begin{table}
\caption{Mutual Information Cuts\label{tab:Mutual-Information-Cuts}}

\begin{centering}
\vphantom{}
\par\end{centering}

\begin{centering}
\begin{tabular}{|c|c|c|}
\cline{2-3} 
\multicolumn{1}{c|}{} & Mean Accuracy & Std. Dev\tabularnewline
\hline 
Best Pairs & 0.6036 & 0.0295\tabularnewline
\hline 
Pairs MI>0 & 0.5518 & 0.0294\tabularnewline
\hline 
Pairs MI>2 & 0.5871 & 0.0289\tabularnewline
\hline 
Pairs MI>4 & 0.5394 & 0.0250\tabularnewline
\hline 
Pairs MI>6 & 0.5340 & 0.0303\tabularnewline
\hline 
\end{tabular}
\par\end{centering}

\begin{centering}
\vphantom{}
\par\end{centering}

A study of the effect of varying the mutual information cut for corpus
pairs. All models were trained on the same parameters (3000 static
features, 240 dynamic features, 1 threshold). The four entries labeled
``Pairs MI>x'' do not have any cuts for rare words or rare pairs,
and show the results of different MI cuts. The entry labeled ``Best
Pairs'' reproduces that from table \ref{tab:Choosing-Significant-Words}:
namely, having three cuts: besides MI>2, it also cuts words that appear
4 or fewer times, and cuts phrases that appear 4 or fewer times. 

\begin{minipage}[t]{1\columnwidth}%
\rule[0.5ex]{1\columnwidth}{1pt}%
\end{minipage}
\end{table}

To improve scores, are the relevant cuts to the rare words, to the
rare phrases, or both? The answer is both, as revealed in table \ref{tab:Cutting-Phrases}.
Of the two, cutting infrequent words seems to provide the greater
benefit.

\begin{table}
\caption{Cutting Phrases\label{tab:Cutting-Phrases}}
\vphantom{}

\begin{centering}
\begin{tabular}{|c|c|c|c|}
\hline 
Word Count Cut & Pair Count Cut & Mean Accuracy & Std. Dev\tabularnewline
\hline 
4 & 4 & 0.6036 & 0.0295\tabularnewline
\hline 
0 & 4 & 0.5871 & 0.0311\tabularnewline
\hline 
0 & 2 & 0.5814 & 0.0351\tabularnewline
\hline 
2 & 2 & 0.5830 & 0.0300\tabularnewline
\hline 
2 & 0 & 0.5908 & 0.0297\tabularnewline
\hline 
4 & 0 & 0.5901 & 0.0309\tabularnewline
\hline 
0 & 0 & 0.5871 & 0.0289\tabularnewline
\hline 
6 & 6 & 0.5711 & 0.0309\tabularnewline
\hline 
\end{tabular}
\par\end{centering}

\begin{centering}
\vphantom{}
\par\end{centering}

A comparison of different word and phrase cuts on the accuracy. All
models are built on the corpus-pairs dataset, and reject pairs with
an MI of 2 or less, and were trained on the same parameters (3000
static features, 240 dynamic features, 1 threshold). The highest scoring
ensemble also rejects words and word-pairs that occur 4 or fewer times.
The other entries explore the effect of cutting fewer or more words
and/or pairs; none do particularly well, not quite approaching the
best result.

\begin{minipage}[t]{1\columnwidth}%
\rule[0.5ex]{1\columnwidth}{1pt}%
\end{minipage}
\end{table}

\subsection*{Static vs. Dynamic Feature Selection}

The use of dynamic feature selection also has a strong effect on both
training times (by reducing the size of the problem) as well as on
the quality of the fit. Table \ref{tab:Static-vs.-Dynamic} shows
the effect of dynamic feature selection on the overall score.

\begin{table}
\caption{Static vs. Dynamic Feature Selection\label{tab:Static-vs.-Dynamic}}

\begin{centering}
\vphantom{}
\par\end{centering}

\begin{centering}
\begin{tabular}{|c|c|c|}
\hline 
Num. Feat. & Static & Dynamic\tabularnewline
\hline 
\hline 
2 &  & 0.5794\tabularnewline
\hline 
3 &  & 0.5983\tabularnewline
\hline 
5 &  & 0.6283\tabularnewline
\hline 
8 &  & 0.6069\tabularnewline
\hline 
12 &  & 0.5981\tabularnewline
\hline 
16 &  & 0.5823\tabularnewline
\hline 
24 & 0.5553 & 0.5691\tabularnewline
\hline 
32 & 0.5580 & 0.5917\tabularnewline
\hline 
50 & 0.5628 & 0.5914\tabularnewline
\hline 
60 &  & 0.5868\tabularnewline
\hline 
75 & 0.5960 & 0.5899\tabularnewline
\hline 
90 & 0.5958 & 0.6227\tabularnewline
\hline 
105 &  & 0.5769\tabularnewline
\hline 
120 & 0.5764 & 0.5772\tabularnewline
\hline 
150 &  & 0.5884\tabularnewline
\hline 
180 & 0.5619 & 0.5906\tabularnewline
\hline 
240 & 0.5526 & 0.6036\tabularnewline
\hline 
300 &  & 0.6028\tabularnewline
\hline 
360 & 0.5436 & \tabularnewline
\hline 
400 &  & 0.5911\tabularnewline
\hline 
500 & 0.5501 & 0.5715\tabularnewline
\hline 
\end{tabular}
\par\end{centering}

\begin{centering}
\vphantom{}
\par\end{centering}

This table shows ensemble averages for the accuracy comparing static
and dynamic feature selection. As usual, the ensemble consists of
100 models; one threshold was used. The dynamic feature-selection
results are identical to those reported in table \ref{tab:Bag-of-Phrases};
they are obtained by statically selecting 3000 features, and then
dynamically selecting the indicated number from those. By contrast,
the static selection process just chooses the indicated number of
features initially, and makes no dynamic selection at all.\\

\includegraphics[width=0.95\columnwidth]{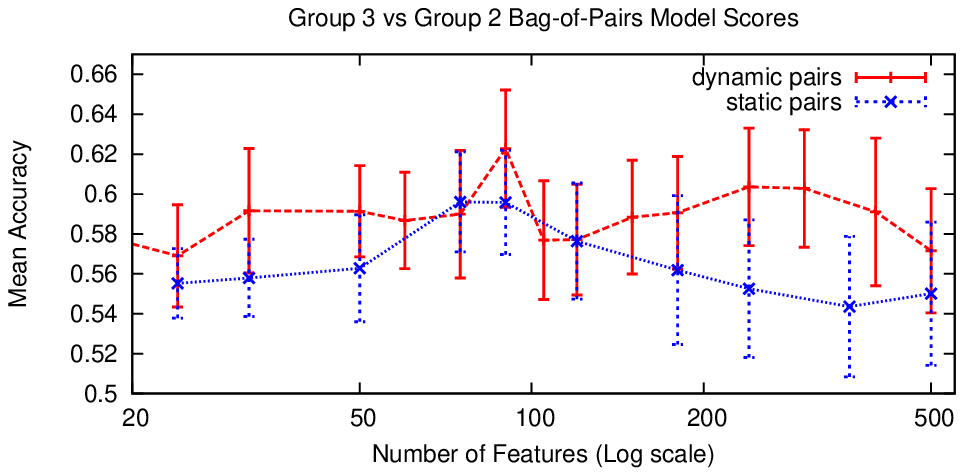}

\end{table}

\subsection*{Evaluation Times}

\begin{figure}
\caption{Train-Test Score Correlation\label{fig:Train-Test-Score-Correlation}}

\includegraphics[width=0.48\columnwidth]{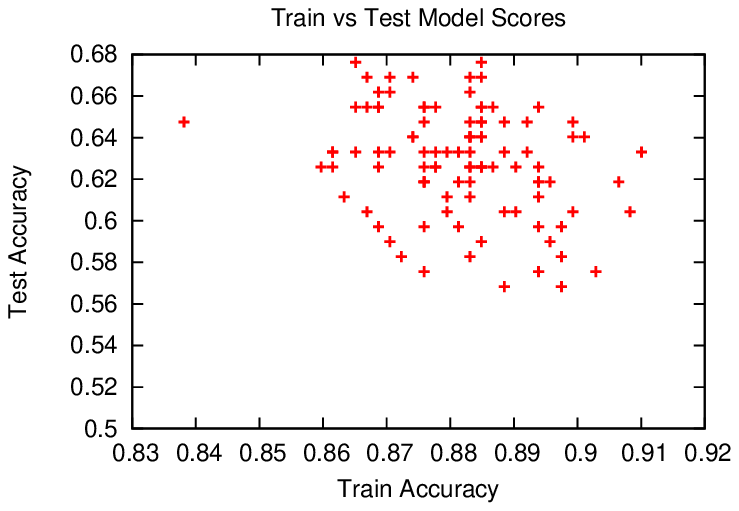}\includegraphics[width=0.48\columnwidth]{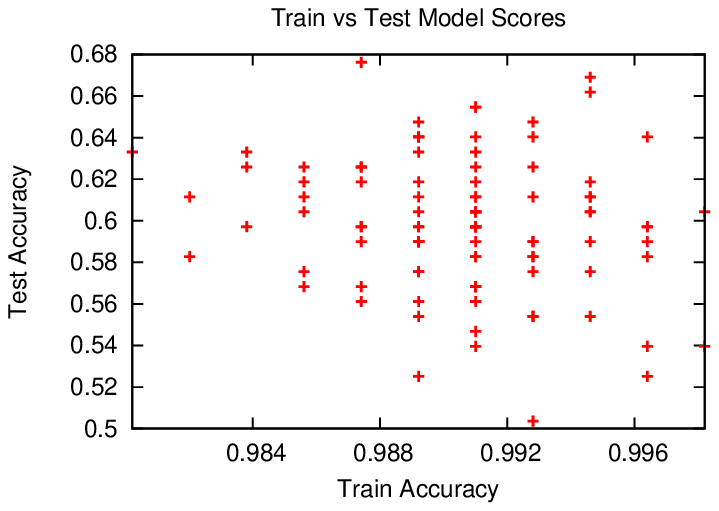}

The above two scatter-plots show the test vs. train scores for 100
models. As should be clear, there is essentially no correlation: a
model that scored well on the training set may or may not do poorly
on the test set, and \emph{vice versa}. The left figure is for 5 dynamical
features, the right figure is for 320. As is clear from the \emph{x}-axis
labels, using more dynamical features helps improve the training score
by a lot; however, the test-score degrades, as noted in table \ref{tab:Bag-of-Phrases}.\\

\begin{minipage}[t]{1\columnwidth}%
\rule[0.5ex]{1\columnwidth}{1pt}%
\end{minipage}
\end{figure}

MOSES is not immune to the effect of over-training: longer training
times result in a better fit to the train dataset, but the resulting
models perform more poorly on the test dataset. In essence, longer
training times allow the system to find quirks in the training set
that are not present in the test set. Once a minimum amount of training
has been done, any correlation between train and test scores disappears;
there is even a vague hint of anti-correlation as shown in figure
\ref{fig:Train-Test-Score-Correlation}. 

Optimum training times are explored in table \ref{tab:Training-Time-Dependence}.
Training times are measured in terms of the number of evaluations
of the scoring function: a single comparison of the model to the training
table counts as one evaluation. The highest score for each row is
marked in bold (\textcolor{magenta}{magenta}); the second-highest
score in italic (\textcolor{cyan}{cyan}). The very highest score,
over the entire table, of 63.68\%, occurs at a training time of 5000
evaluations and 90 dynamical features (indicated in bold sans-serif).
Most of the high scores occur when 9000 or fewer training evaluations
are performed. The exceptions occur when the number of dynamical features
is extremely small: this suggests that the model builder is starved
for features at this point, and must iterate over many trials before
finding the appropriate features. 

\begin{table}
\caption{Training Time Dependence\label{tab:Training-Time-Dependence}}

\begin{centering}
\vphantom{}
\par\end{centering}

\begin{centering}
\begin{sideways}
\begin{tabular}{|>{\centering}p{0.06\paperwidth}|c|c|c|c|c|c|>{\centering}p{0.06\paperwidth}|>{\centering}p{0.06\paperwidth}|>{\centering}p{0.06\paperwidth}|>{\centering}p{0.06\paperwidth}|}
\hline 
Dyn. Feat & 3K Eval & 5K Eval & 9K Eval & 15K Eval & 27K Eval & 45K Eval & 70K Eval & 110K Eval & 160K Eval & 320K Eval\tabularnewline
\hline 
\hline 
3 & 0.5914 & 0.5965 & \textcolor{black}{0.5967} & 0.5955 & 0.5922 & 0.5937 & 0.5959 & \textit{\textcolor{cyan}{0.5973}} & \textbf{\textcolor{magenta}{0.5983}} & 0.5965\tabularnewline
\hline 
5 & 0.5959 & 0.6192 & 0.6271 & \textit{\textcolor{cyan}{0.6288}} & 0.6281 & 0.6268 & 0.6265 & 0.6281 & \textcolor{black}{0.6283} & \textbf{\textcolor{magenta}{0.6301}}\tabularnewline
\hline 
8 & \textit{\textcolor{cyan}{0.6185}} & \textbf{\textcolor{magenta}{0.6229}} & \textcolor{black}{0.6140} & 0.6122 & 0.6099 & 0.6066 & 0.6062 & 0.6068 & 0.6069 & 0.6055\tabularnewline
\hline 
12 & 0.5871 & 0.6049 & \textit{\textcolor{cyan}{0.6088}} & \textbf{\textcolor{magenta}{0.6091}} & \textcolor{black}{0.6055} & 0.6022 & 0.6002 & 0.5995 & 0.5981 & 0.5942\tabularnewline
\hline 
16 & 0.5563 & 0.5796 & \textbf{\textcolor{magenta}{0.5957}} & \textit{\textcolor{cyan}{0.5932}} & \textcolor{black}{0.5925} & 0.5886 & 0.5855 & 0.5847 & 0.5823 & \tabularnewline
\hline 
24 & 0.5678 & 0.5678 & \textbf{\textcolor{magenta}{0.5765}} & 0.5671 & 0.5681 & 0.5682 & 0.5684 & \textit{\textcolor{cyan}{0.5692}} & 0.5691 & 0.5681\tabularnewline
\hline 
32 & \textbf{\textcolor{magenta}{0.5965}} & 0.5720 & 0.5732 & 0.5923 & \textit{\textcolor{cyan}{0.5945}} & \textcolor{black}{0.5929} & 0.5929 & 0.5927 & 0.5917 & 0.5918\tabularnewline
\hline 
40 & 0.6214 & \textit{\textcolor{cyan}{0.6240}} & \textbf{\textcolor{magenta}{0.6288}} & \textcolor{black}{0.6221} & \textcolor{black}{0.6141} & 0.6139 & 0.6140 & 0.6136 & 0.6132 & \tabularnewline
\hline 
50 & \textbf{\textcolor{magenta}{0.6273}} & \textit{\textcolor{cyan}{0.6145}} & 0.5924 & \textcolor{black}{0.5971} & \textcolor{black}{0.5935} & 0.5924 & 0.5927 & 0.5919 & 0.5914 & 0.5899\tabularnewline
\hline 
60 & \textit{\textcolor{cyan}{0.6244}} & \textbf{\textcolor{magenta}{0.6277}} & \textcolor{black}{0.6060} & \textcolor{black}{0.5876} & 0.5868 & \textcolor{black}{0.5872} & 0.5868 & 0.5871 & 0.5868 & \tabularnewline
\hline 
75 & \textit{\textcolor{cyan}{0.6240}} & \textbf{\textcolor{magenta}{0.6245}} & \textcolor{black}{0.5982} & \textcolor{black}{0.5921} & 0.5893 & 0.5894 & 0.5895 & \textcolor{black}{0.5902} & 0.5899 & 0.5894\tabularnewline
\hline 
90 & 0.6143 & \textsf{\textbf{\textcolor{green}{0.6368}}} & 0.6204 & 0.6201 & 0.6227 & \textit{\textcolor{cyan}{0.6231}} & 0.6224 & 0.6222 & 0.6227 & 0.6212\tabularnewline
\hline 
105 & 0.5813 & \textbf{\textcolor{magenta}{0.6004}} & \textit{\textcolor{cyan}{0.5863}} & 0.5750 & 0.5768 & 0.5765 & 0.5768 & 0.5765 & \textcolor{black}{0.5769} & \tabularnewline
\hline 
120 & \textit{\textcolor{cyan}{0.6055}} & \textbf{\textcolor{magenta}{0.6112}} & \textcolor{black}{0.6045} & \textcolor{black}{0.5799} & 0.5778 & 0.5781 & \textcolor{black}{0.5783} & 0.5776 & 0.5772 & 0.5768\tabularnewline
\hline 
150 & 0.5878 & \textit{\textcolor{cyan}{0.6168}} & \textbf{\textcolor{magenta}{0.6200}} & \textcolor{black}{0.5951} & 0.5897 & \textcolor{black}{0.5904} & 0.5893 & 0.5886 & 0.5884 & \tabularnewline
\hline 
180 & 0.5839 & \textit{\textcolor{cyan}{0.6063}} & \textbf{\textcolor{magenta}{0.6070}} & 0.6006 & 0.5919 & 0.5913 & 0.5908 & 0.5901 & 0.5906 & 0.5898\tabularnewline
\hline 
240 & 0.5891 & 0.6040 & \textbf{\textcolor{magenta}{0.6145}} & \textit{\textcolor{cyan}{0.6095}} & 0.6028 & 0.6040 & 0.6037 & 0.6035 & 0.6036 & 0.6024\tabularnewline
\hline 
300 & 0.5874 & 0.5890 & \textbf{\textcolor{magenta}{0.6170}} & \textit{\textcolor{cyan}{0.6067}} & 0.6045 & 0.6033 & 0.6029 & 0.6029 & 0.6028 & 0.6027\tabularnewline
\hline 
400 &  &  &  &  &  &  &  &  & 0.5911 & \tabularnewline
\hline 
500 &  &  &  &  &  &  &  &  & 0.5715 & \tabularnewline
\hline 
\end{tabular}
\end{sideways}
\par\end{centering}

\centering{}\vphantom{}
\end{table}

\subsection*{Voting}

The concept of the ensemble replaces a significant random variation
in the accuracy of a single representation with a more trustworthy
average accuracy across multiple representations. This does not imply
that the accuracy of the ensemble model is equal to the average accuracy
of the representations in the ensemble. When multiple representations
are allowed to vote for a final classification, the accuracy of the
classifier usually increases\cite{Opitz1999}. This section explores
how the accuracy of a model depends on the number of representations
voting in the model.

\begin{figure}
\begin{centering}
\caption{Voting Ensemble\label{fig:Ensemble-Averages}}
\vphantom{}
\par\end{centering}

\begin{centering}
\includegraphics[width=0.95\columnwidth]{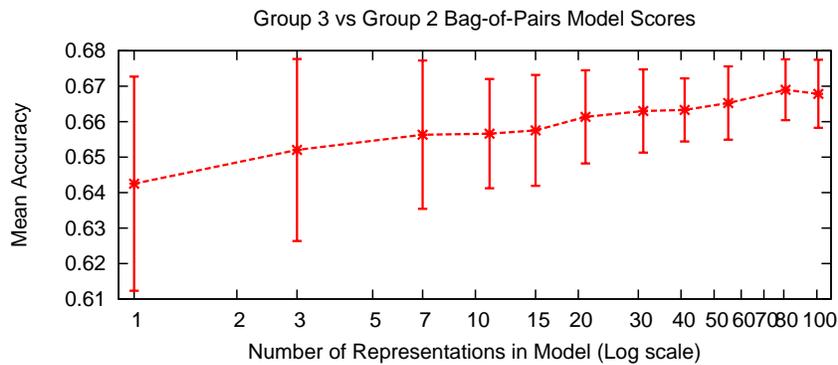}
\par\end{centering}

\begin{centering}
\vphantom{}
\par\end{centering}

Above is a graph of the voting model accuracy for the va32 dataset
(750 significant features, -m=9000 training time). Raw data taken
from table \ref{tab:Ensemble-Averages}\\

Far left is the best result for models containing a single representation:
\emph{i.e.} poses was trained 100 times, on the same parameters, varying
only the initial random seed.  The average accuracy was 64.25\%  The
error bars show the variation among these 100 models: some scored
as high as 69\%, some were down in the mid-upper 50's. \\

The next point over shows the results for a model containing N=3 representations.
 That is, 3 different random seeds are used to create 3 representations.
These are placed in a model, and these then 'vote' for the most likely
classification (the cohort that gets 2 or more votes 'wins').  This
process is repeated 100 times (same parameters, thus 300 different
random seeds). The average accuracy of 100 N=3 models is 65.63\%  The
error bars again show the variation among these 100 models: the best
scoring model hit 69\%, the worst-scoring had a low of 61\%\\

The rightmost point is for a model holding 101 representations.  The
average accuracy (of 100, etc.) is 66.53\%  the best scoring is 69\%.
The worst-scoring is 63\%.  Notice that the best scores are always
pegged at 69\%  The ensemble seems to merely trim away the outliers
with the bad scores.

\begin{minipage}[t]{1\columnwidth}%
\rule[0.5ex]{1\columnwidth}{1pt}%
\end{minipage}
\end{figure}

\begin{figure}

\caption{Distribution of Voting Models\label{fig:Distribution-of-Voting}}

\includegraphics[width=0.95\columnwidth]{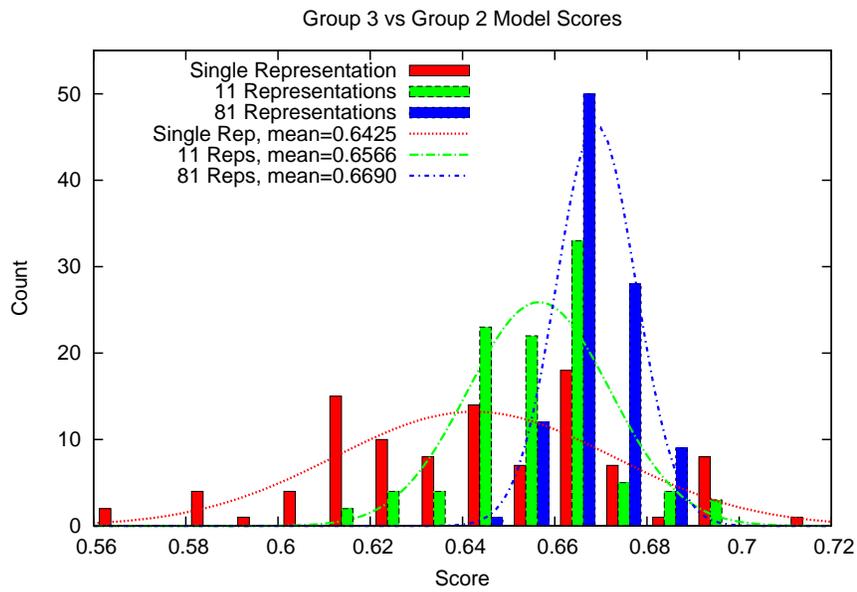}

This figure demonstrates typical distributions taken from figure \ref{fig:Ensemble-Averages};
the means and widths of the Gaussians shown here are exactly those
of table \ref{tab:Ensemble-Averages}.

\begin{minipage}[t]{1\columnwidth}%
\rule[0.5ex]{1\columnwidth}{1pt}%
\end{minipage}
\end{figure}

The results here are reported in the same fashion as before, except
that now, each model contains N representations, instead of just one
representation. In essence, there is now an ensemble of ensembles:
although a model may consist of N representations, we still explore
the average accuracy taken over 100 models. The raw data are presented
in \ref{tab:Ensemble-Averages} and graphed in \ref{fig:Ensemble-Averages}.
Typical cross-sections are shown in \ref{fig:Distribution-of-Voting}.

\begin{table}
\begin{centering}
\caption{Voting Ensemble Averages\label{tab:Ensemble-Averages}}

\par\end{centering}

\begin{centering}
\vphantom{}
\par\end{centering}

\begin{centering}
\begin{tabular}{|c|c|c|}
\hline 
Model Size & Mean & Std. Dev.\tabularnewline
\hline 
\hline 
1 & 0.6425 & 0.0302\tabularnewline
\hline 
3 & 0.6520 & 0.0257\tabularnewline
\hline 
7 & 0.6563 & 0.0209\tabularnewline
\hline 
11 & 0.6566 & 0.0154\tabularnewline
\hline 
15 & 0.6576 & 0.0156\tabularnewline
\hline 
21 & 0.6614 & 0.0131\tabularnewline
\hline 
31 & 0.6630 & 0.0117\tabularnewline
\hline 
41 & 0.6633 & 0.0089\tabularnewline
\hline 
55 & 0.6653 & 0.0104\tabularnewline
\hline 
81 & 0.6690 & 0.0086\tabularnewline
\hline 
101 & 0.6678 & 0.0096\tabularnewline
\hline 
\end{tabular}
\par\end{centering}

\begin{centering}
\vphantom{}
\par\end{centering}

Voting ensemble results. Each model consists of N='Model Size' representations,
with a majority vote determining how the model classifies. The mean
and standard deviation are obtained by averaging over 100 different
models, built by varying the initial random number seed of the machine
learning system. Note that the standard deviation of the N=1 model
is comparable to that of \ref{fig:Ensemble-Average}; as N increases,
the score improves, and the variation shrinks sharply.\\

All models were word-pair models, with the usual word-pair cuts (mi>2,
and all single-words and word pairs that appear less than 4 times
discarded). Word pairs were considered only if they one of the two
words were one of the top 750 most score-correlated single words (this
differs from the other reported word-pair results, where 371 words
were used to create pairs; thus scores are not immediately comparable).
The run-time dynamical feature count was set to 90, and a total of
9000 training evaluations were performed. These parameters are more
or less identical to those discussed for much of this paper, and,
for the N=1 case, correspond to the highest score seen. \\

This table is graphed in figure \ref{fig:Ensemble-Averages}.

\begin{minipage}[t]{1\columnwidth}%
\rule[0.5ex]{1\columnwidth}{1pt}%
\end{minipage}
\end{table}

\begin{figure}
\caption{Voting Blocks\label{fig:Voting-Blocks}}

\begin{centering}
\vphantom{}
\par\end{centering}

\includegraphics[width=0.95\columnwidth]{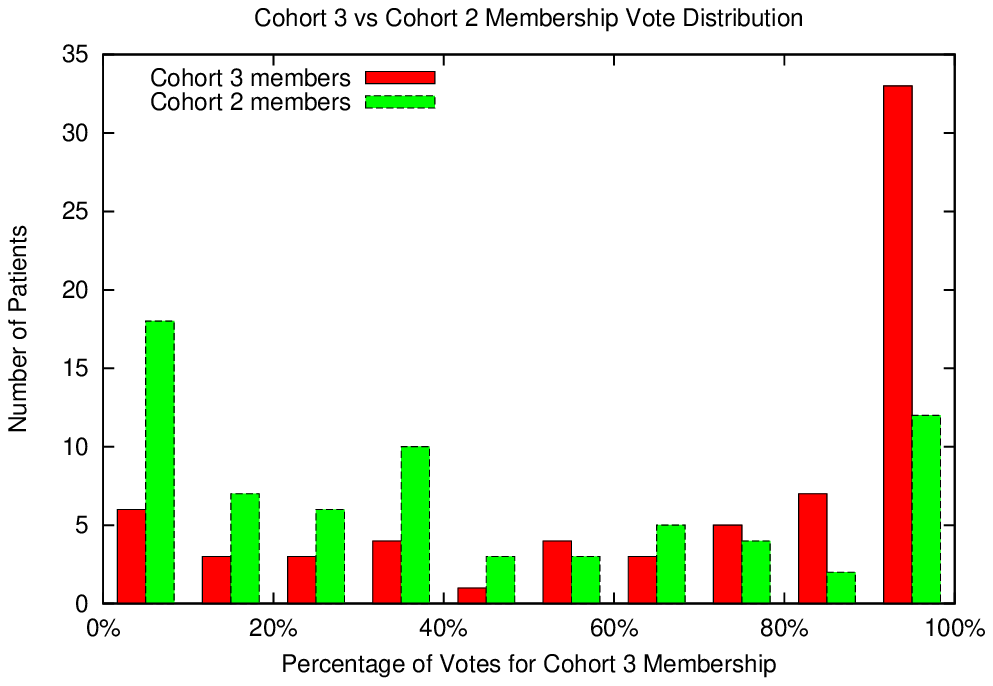}

\begin{centering}
\vphantom{}
\par\end{centering}

This figure shows a model comprising N=101 representations, voted
to classify individual patients. Vote counts were divided into 10
bins; the number of patients receiving that number of votes is shown.
Thus, the tall green bar on the far left indicates that 18 patients
from group 2 received 10\% or fewer votes; these patients are correctly
classified by the voting. By contrast, the red bar on the far left
indicates that 6 patients from group 3 received 10\% or fewer votes;
these patients are misclassified by voting. Indeed, all red bars to
the left of the 50\% mark, and all green bars to the right of the
50\% mark indicate misclassified patients.\\

Just as in the rest of this paper, this shows the performance of the
classifier on the test set, using 5-fold validation. Different models,
created with different random seeds, show a very nearly identical
vote distribution. 

\begin{minipage}[t]{1\columnwidth}%
\rule[0.5ex]{1\columnwidth}{1pt}%
\end{minipage}
\end{figure}

Additional insight can be gained by examining how the representations
voted for individual patients. This is shown in figure \ref{fig:Voting-Blocks}.
Any given patient can receive anywhere from 0\% to 100\% of the votes.
A vote 'for' indicates the patient belongs to group 3, a vote 'against'
indicates that the patient belongs to group 2. Thus, those receiving
less than 50\% of the vote are classified as group 2; those receiving
more are classified as group 3. The graph then shows the fraction
of votes received, versus the known a priori patient cohort membership.
Ideally, a 100\% accurate classifier would always give more than 50\%
of the votes to group 3 members, and always less than 50\% of the
votes to group 2 members. The fact that the classifier is sometimes
wrong is readily apparent in the graph.

A notable feature of the graph is that it is not symmetrical: that
is, the red bars are not a mirror image of the green bars. Of particular
interest is that the classifier is overall quite confident in its
classification of group 3 patients (this is the psychiatric group);
this can be seen in the tall bar on the right-hand side of the graph.
That is, given a patient from group 3, the classifier can correctly
classify the patient with good accuracy and high confidence. This
is not at all the case from group 2, the suicide cohort: here, the
classifier is clearly less accurate, and more tentative in its assignment.
This can be seen in that the left-most green bar is not that tall,
and that the rightmost green bar is not very small, as one might have
hoped. In essence, the classifier is good at recognizing the psychiatric
patients; but the suicidal patients, not so much.

\subsection*{Future Work}

The current datasets were balanced in the number of patients; but
suicide risk is small in the general population. A classification
system deployed on a large scale would need to be able to cope with
this, to pull the proverbial needle from the haystack. Thus, for future
Durkheim project work, it seems most appropriate to optimize for recall,
rather than accuracy. The recall rate of a classifier measures how
well the classifier is able to identify true-positives, possibly at
the expense of a high false-positive rate. The core presumption here
is that one would rather be 'safe than sorry': to over-asses suicide
risk, so as not to miss any true-positives. For such general-population
classifiers, it seems that the best approach would be to maximize
the recall rate, while clamping the false-positive rate below a reasonable
level. Another alternative would be to maximize the $F_{2}$-score,
which is a weighted harmonic mean of the recall and precision of the
model. 

In the dataset, words such as ``worthlessness'' appear far more
often in group 2 than in the other groups. The word ``despondent''
appears only in group 2, and there are highly elevated counts of the
words ``agitation'' and ``aid'' in this group. By contrast, some
words are remarkable by their absence: the words ``crying'' and
``aggravating'' are absent or nearly absent in group 2, and appear
primarily in group 3. This may be due to a difference in the psychological
coping abilities and strategies in these two groups, although it may
also reflect the small sample size. In the same vein, ``obesity''
appears half as often in group 2 as in group 3: perhaps an eating
disorder is one way of coping? Without a fuller context, such as the
standard approach of corpus linguistics, it is hard to tell.

Given these observations on word counts, a promising avenue for future
research would be to further explore the corpus linguistics-inspired
approach. Rather than creating a bag-of-words, the core idea would
be to create a more refined ``bag-of-phrases'', with phrases constructed
not only from nearest neighbors, but perhaps derived from, or incorporating
syntactic information, such as part-of-speech tags, dependency tags
from a dependency parse, or even semantic information, such as WordNet
lexical tags\cite{Mihalcea2005,Mihalcea2007}. A separate challenge
in the dataset is the presence of question-answer constructions, with
the answer being relevant to evaluating psychological state, whereas
the the question is worded with psychologically significant words
that would confuse a bag-of-words/bag-of-phrases classifier. Techniques
from anaphora resolution algorithms or perhaps tricks from question-answering
systems might be applicable to disambiguate the intended meaning.

\section*{Conclusion}

Training classifiers to distinguish the three groups of patients is
a straightforward task. Given the relatively small dataset size, it
was also easy to train these classifiers to be ``over-fit'': to
perform very well on the training set, sometimes achieving a perfect
score, but scoring rather poorly on the test set. Accuracies up to
67\% were obtained for ensemble averages of 100 models, trained on
the best parameter choices, with individual model accuracies rising
as high as 69\%. 

Finding the best models is an arduous task. To evaluate an ensemble
of 100 models with 5-fold cross-validation requires a total of 500
models to be trained; this can take days of wall-clock time, as individual
models require anywhere from a few minutes to a decent fraction of
an hour to train. In order to obtain a good fit, several training
parameters must be explored: the thresholding of word-counts into
bins, and the run-time dynamical feature-selection size. These parameters
must be tuned individually for different data sets; they are adjusted
to best bring a view of the dataset into sharp focus.

The most interesting result is that word-pairs can be used to build
more accurate models than single words alone. However, in order for
this to work well, a number of data cuts must be applied: word pairs
with low mutual information scores should be discarded; infrequently
occurring pairs and words should be discarded, and, most important
of all, word-pairs that don't contain 'significant' words should be
discarded as well.

\subsection*{Acknowledgments}

This work was partially funded by DARPA grant \# N66001-11-C-4006,
and IARPA grant \# N10PC20221. The author would like to thank Chris
Poulin and Ben Goertzel for being encouraging and patient as this
work was carried out. The system described here is a part of the Durkheim
Project \url{http://durkheimproject.org/}.

\bibliographystyle{plain}
\bibliography{local}

\end{document}